\documentclass{article}

\usepackage[preprint]{neurips_2026}

\usepackage[utf8]{inputenc} 
\usepackage[T1]{fontenc}    
\usepackage{hyperref}       
\usepackage{url}            
\usepackage{booktabs}       
\usepackage{amsfonts}       
\usepackage{nicefrac}       
\usepackage{microtype}      
\usepackage{xcolor}         

\usepackage{graphicx}
\usepackage{multirow}
\usepackage{arydshln}
\usepackage[most]{tcolorbox}
\usepackage{inconsolata}
\tcbuselibrary{listings}
\usepackage{amssymb}

\usepackage{algorithm}
\usepackage[noend]{algpseudocode}
\algrenewcommand\algorithmicrequire{\textbf{Input:}}
\algrenewcommand\algorithmicensure{\textbf{Output:}}

\usepackage{threeparttable}

\usepackage{wrapfig}

\newtcolorbox{promptbox}[1]{
    listing only,
    colback=gray!10,
    colframe=gray!80!black,
    colbacktitle=gray!25!black,
    coltitle=white,
    fonttitle=\bfseries\sffamily,
    fontupper=\ttfamily\small,
    arc=5pt,
    title=#1,
    enhanced,
    breakable,
    listing options={
        basicstyle=\ttfamily,
        breaklines=true,
        breakatwhitespace=false,
        columns=fullflexible,
        keepspaces=true
    },
    attach boxed title to top left={yshift=-2mm, xshift=2mm},
    boxed title style={sharp corners=south, size=small}
}

\title{WebChallenger: A Reliable and Efficient Generalist Web Agent}

\author{%
  Jayoo Hwang\thanks{Corresponding author.} \\
  ML Collective \\
  \texttt{jayoohm350@gmail.com} \\
  \And
  Xiaowen Zhang \\
  longsurf.ai \\
  \texttt{sean@longsurf.ai} \\
  \And
  Vedant Padwal \\
  Independent \\
  \texttt{vedantpadwalinfi@gmail.com } \\
}

\begin{document}

\maketitle

\begin{abstract}
Autonomous web navigation remains challenging for LLM agents, and the strongest
generalist systems rely on proprietary reasoning models whose inference cost is
prohibitive for the repetitive tasks where such agents would be most useful.
We argue this gap stems not from insufficient model capability but from agent
architectures that fail to replicate three human cognitive advantages: selective
attention to relevant page regions, persistent memory of website structure, and
procedural fluency with common interaction patterns. We introduce WebChallenger, a web agent
framework that addresses each gap through architecture design rather than model
scale, built around PageMem: a structured page representation
deterministically constructed from the DOM that exposes each page as a hierarchy
of semantic sections with short summaries. On this shared substrate we build
three mechanisms that mirror the three cognitive advantages: a
divide-and-conquer observation pipeline that lets the agent skim section
summaries and extract details only from task-relevant regions; a lightweight
exploration and memory system that traverses each website once to build a
reusable map of pages and element behaviors; and compound action workflows that
collapse common multi-step interactions into single agent actions, handling
partial state changes automatically. Because all three operate over PageMem,
the framework generalizes across websites without site-specific adapters.
Using off-the-shelf open-weight models without
fine-tuning, our system achieves 56.3\% on WebArena, 48.7\% on VisualWebArena,
51.0\% on Online-Mind2Web, and 70.9\% on WorkArena, approaching frontier
proprietary systems at a fraction of the cost. Our code is released at this \href{https://github.com/jayoohwang1/webchallenger}{URL}.
\end{abstract}

\section{Introduction}
\label{introduction}

\begin{quote}
    \itshape
    ``I touch the future. I teach'' 
    \par --- Christa McAuliffe
\end{quote}

\begin{figure}  
\includegraphics[width=12cm]{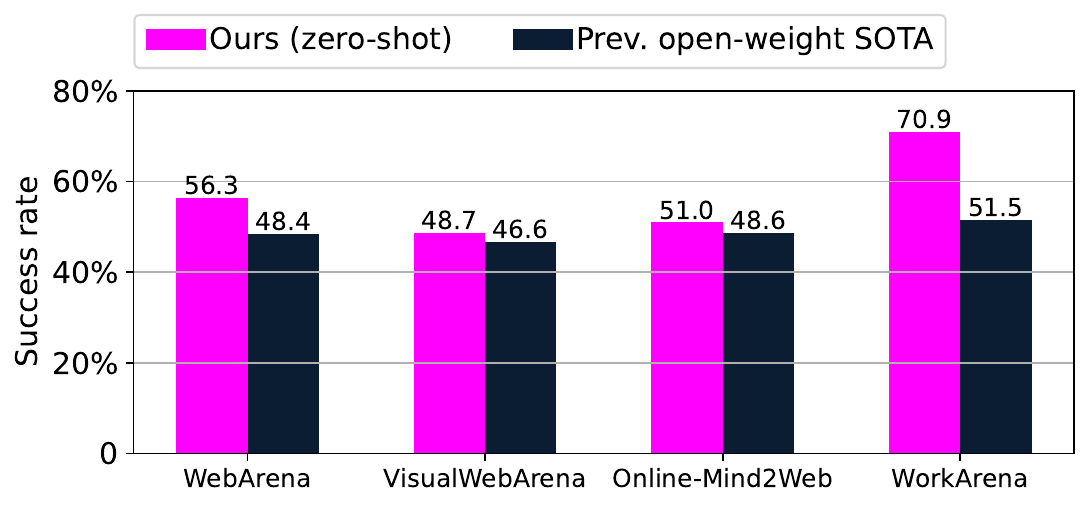}
\caption{Benchmark results. WebChallenger sets new state-of-the-art performance among agents using open models across four web navigation benchmarks. Our results were obtained with far less compute than the baselines which either used finetuning or larger models, demonstrating that \textbf{scaffolding alone can drastically improve web agent performance.}}
\label{figure_1}
\end{figure}

Autonomous web navigation has long been a goal of AI research \citep{10.1145/267658.267666}: the web is one of the most complex
interactive environments available, and navigating it autonomously has broad practical implications,
from automating repetitive knowledge work to serving as a testbed for general-purpose agent
capabilities. Recent advances in large language models and vision-language models have driven rapid
progress on computer-using agents \citep{marino2025computeruse}, yet even the strongest LLM agents remain below human
performance on realistic, long-horizon web tasks \citep{jang2026odysseysbenchmarkingwebagents, miyai2025webchorearenaevaluatingwebbrowsing}. Additionally, the best generalist agents rely on
proprietary reasoning models whose inference cost is prohibitive for the repetitive work where 
agents would be desirable.

This gap echoes Moravec's paradox \citep{moravec1988mind, su2025computeruse}: browsing the web is effortless for humans yet remarkably
difficult for AI models that excel at mathematics and code generation. We argue that this
difficulty stems not from a lack of web knowledge in current models, but from a mismatch between
how agent frameworks present the web environment and how it needs to be processed. Specifically,
humans bring three cognitive advantages to web navigation that current agent architectures fail to
replicate. First, \textit{selective attention}: humans focus on relevant regions of a page while
ignoring the rest \citep{10.1007/978-3-031-28238-6_5}, whereas LLM agents ingest entire pages as flat token sequences, diluting
relevant information in irrelevant context. Second, \textit{persistent memory}: humans memorize the
layout and functionality of websites they have used before, while LLM agents approach each session
with no prior environmental knowledge. Third, \textit{procedural fluency}: humans internalize
reusable routines for common interaction patterns (e.g., searching, selecting from a dropdown, filling
a form) that execute as cohesive sequences without deliberate reasoning at each step, while LLM
agents must re-observe and re-reason over the full page state for every atomic action.

In this work, we show that these three human advantages can be realized through agent architecture
design rather than model scale or training. Implementing them in a way that generalizes across
websites without site-specific adapters requires a shared abstraction the agent
can reason over uniformly. We introduce \textbf{PageMem}, a structured page representation
deterministically constructed from the DOM that exposes each page as a hierarchy of semantic
sections with short summaries: a representation the agent can skim like a table of contents,
expand selectively for detail, and dispatch to specialized workflows by section type. On this
substrate we build three mechanisms that mirror the three cognitive advantages above.

A \textbf{divide-and-conquer observation} pipeline lets the agent skim PageMem's section summaries,
select task-relevant regions, and extract details only from those regions, producing
information-dense observations without processing entire pages.

A lightweight \textbf{exploration and memory} system traverses new websites before task execution,
assembling a persistent collection of PageMems that records pages, navigation paths, and
interactive element behaviors.

\textbf{Compound action workflows} implement site-agnostic routines for common interaction
patterns such as searching, menu selection, and form submission. Dispatched by section type, these
workflows collapse multi-step processes into single agent actions and automatically surface
partial state changes (such as a dropdown expanding) without requiring the agent to
reprocess the full page.

Decomposing observation and decision-making into focused sub-prompts in this way allows our
framework to extract strong performance from small, locally-run models that would struggle with
the monolithic prompts used by most existing agent frameworks. Using an off-the-shelf 32B LLM and
a 7B VLM without any fine-tuning, our system achieves 56.3\% on WebArena
\citep{zhou2024webarena}, 48.7\% on VisualWebArena \citep{koh2024visualwebarena}, 51.0\% on
Online-Mind2Web \citep{xue2025onlinemind2web}, and 70.9\% on WorkArena
\citep{drouin2024workarenacapablewebagents} --- state-of-the-art results among open-weight models
of comparable scale, and approaching frontier proprietary systems at a fraction of the inference cost.
These results indicate that current LLMs already possess sufficient reasoning ability for many
web tasks; what they lack is the right scaffolding around observation, memory, and action to use
it effectively.

\section{Method}
\subsection{Problem Formulation}

We frame web navigation as a sequential decision process in which an agent interacts with a web browser to complete a natural-language task. A task is a tuple $\tau = (I, u_0)$ consisting of an instruction $I$ and a starting URL $u_0$, which determines the initial website $w_0$ from a set $\mathcal{W}$ of target websites. At each timestep $t$, the agent receives an observation $o_t$, maintains a compact history $h_t$ of prior interactions, and selects an action $a_t$ from a candidate set $\mathcal{A}_t$.

A standard LLM web agent implements this loop as $a_t = \pi(o_t, h_t)$: a single model call that maps a raw observation — typically a full accessibility tree or screenshot — and an interaction history to the next atomic browser action. Our system departs from this template with four novel components.

\paragraph{A structured page representation.} Rather than exposing the raw DOM or accessibility tree, we introduce \textbf{PageMem}, a structured representation $p$ deterministically constructed from the DOM. Each PageMem contains an ordered list of \textbf{PageSections} $\{s_1, \ldots, s_n\}$ corresponding to semantic regions of the page, and each PageSection contains a set of interactive \textbf{Elements}. PageSections carry model-generated summaries alongside DOM-derived attributes, and serve as the shared substrate on which the observation pipeline, memory, and action workflows all operate. This abstract substrate is what allows the rest of the system to remain site-agnostic. PageMem is defined in detail in \S\ref{sec:pagemem}.

\paragraph{Persistent memory from offline exploration.} Before any task is attempted, an offline exploration phase traverses each website $w \in \mathcal{W}$ and builds a \textbf{WebsiteMem} $\mathcal{M}_w$: a persistent collection of PageMems indexed by URL, together with information about page templates and element behaviors discovered during exploration. At task start the agent may select a set of bookmarks $B_\tau \subseteq \mathcal{M}_{w_0}$ that remain available as navigation targets throughout the task. WebsiteMem is constructed once per site and reused across all subsequent tasks. Exploration and memory are detailed in \S\ref{sec:exploration}.

\paragraph{A multi-stage observation pipeline.} Rather than producing $o_t$ by serializing the full page, we decompose observation into three stages over the current PageMem $p_t$: the agent first selects a subset of sections whose summaries appear relevant to the task, then extracts task-relevant details from the full content of each selected section, and finally synthesizes the extractions into a task-focused page summary $\hat{o}_t$. The pipeline is defined in \S\ref{sec:observation}.

\paragraph{Compound actions with workflows.} A timestep in our system corresponds to one \textit{high-level} agent action, which may execute multiple browser operations. Single-step actions (clicking a link, navigating to a URL) cause a page transition and advance the loop directly. Compound actions (dropdown selection, form submission, search) invoke a \textbf{workflow} $\omega(a_t)$ — a fixed sequence of additional LLM sub-calls and browser operations that handles intermediate partial state changes, such as a dropdown expanding or form fields being filled one at a time, before returning control to the top-level loop. The action system is detailed in \S\ref{sec:actions}.

\subsubsection{System overview.} Given a task $\tau = (I, u_0)$, the agent retrieves the WebsiteMem $\mathcal{M}_{w_0}$ built during offline exploration and optionally selects bookmarks $B_\tau$. At each timestep $t$, it (i) retrieves or constructs the PageMem $p_t$ for the current page; (ii) applies the observation pipeline to produce $\hat{o}_t$; and (iii) selects an action $a_t \in \mathcal{A}_t$, which executes either as a direct browser operation or through a workflow $\omega(a_t)$. The loop terminates when the agent selects an end-task action and verifies completion, or when a step budget is exhausted. The agent inference algorithm is provided in Appendix~\ref{app:agent-loop}.

\begin{figure} 
\begin{center}
\centerline{\includegraphics[width=\columnwidth]{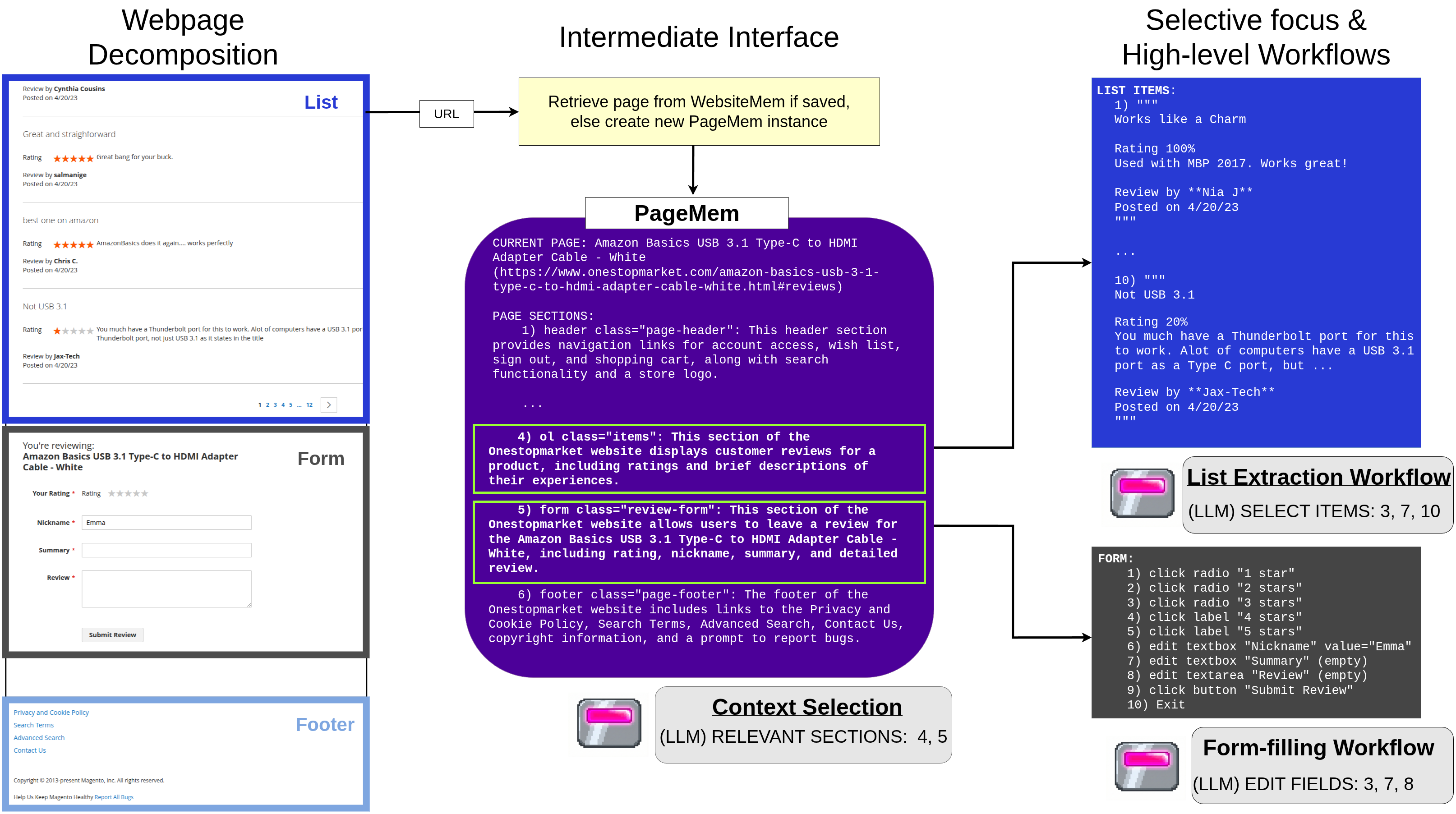}}
\vskip -0.1in
\caption{Overview of WebChallenger. (left) Each webpage is decomposed along the DOM into sections that correspond to semantic regions of the page. (middle) These sections are indexed by short summaries to form a PageMem, a structured page representation cached in per-website memory. The agent skims these summaries and expands only the task-relevant sections for detailed processing. (right) Specialized multi-step workflows are executed based on section type.}
\label{fig:main}
\end{center}
\vskip -0.37in
\end{figure}

\subsection{PageMem}
\label{sec:pagemem}

PageMem is an abstract page representation deterministically constructed from the DOM that serves as the common interface shared by the exploration (\S\ref{sec:exploration}), observation (\S\ref{sec:observation}), and action (\S\ref{sec:actions}) components. It exposes a semantic, chunked view of a page while preserving the selectors needed for direct browser control, allowing higher-level components to operate on abstract objects without site-specific adapters.

\paragraph{Hierarchy.}
PageMem is organized in four levels. A \emph{WebsiteMem} $\mathcal{M}_w$ contains all PageMems and elements encountered on a website $w$. A \emph{PageMem} $p$ corresponds to a single page and holds a title, URL, ordered list of sections $(s_1, \ldots, s_n)$, and a page-level summary. A \emph{PageSection} $s_i$ represents a subregion of the page (e.g., navigation bar, product listing, review form) and maps to a sub-tree of the DOM. Each section carries DOM-derived state attributes (e.g., tag, class, bounding box, contained elements) and variable metadata (e.g., summary, extracted details). An \emph{Element} $e$ represents a single interactive widget, and carries DOM attributes to enable selector construction as well as metadata such as the element's current value, clicked status, and dropdown elements. The PageMem data structure acts as the central hub where all agent-related information about a page is stored, flexibly facilitating the implementation of precise context-engineering for web agents.

\paragraph{Construction.}
PageSections are produced by recursively splitting the DOM tree, terminating at nodes that either fall below a size threshold or match a grouping tag (\texttt{form}, \texttt{ul}, \texttt{li}, \texttt{table}, \texttt{section}, etc.); sibling nodes sharing tag and class are grouped into a single \emph{list section}. Clickable elements are identified using heuristics adapted from the BrowserUse library~\citep{browser-use} and assigned to their ancestor section. Finally, we prompt an LLM or VLM to provide a general one-sentence summary for each section and the overall page. Normal sections are size-bounded so their full content fits in a single LLM call; list sections are unbounded and represented at a higher level of abstraction as a sequence of uniform sub-sections, one per item. Full details and construction algorithm are given in Appendix~\ref{app:pagemem}.

\subsection{Exploration and Memory}
\label{sec:exploration}

Before any task is attempted, an offline exploration phase traverses each target website $w \in \mathcal{W}$ and produces the WebsiteMem $\mathcal{M}_w$ used at inference. Exploration is fully deterministic: it requires no LLM guidance, task demonstrations, or external resources. Compared to tree-search methods that expand during execution or skill-learning approaches that improve only after accumulating task experience, our approach amortizes environmental knowledge upfront and makes it available from the first task at a fixed, one-time cost. We describe exploration here and provide details in Appendix~\ref{app:exploration}.

\paragraph{Traversal.}
Starting from the homepage of a website, we explore all unique clickable elements on the page in order. If a page contains many repeated elements with the same structure (such as a list or table of results), then we only explore elements contained within one item/row of the list/table for efficiency.
We skip exploring elements that have already been explored on the current website. An element is explored by clicking it and recording the state-transition it induces. If clicking results in navigation to an \textit{unexplored page}, the URL of the new page is added to the exploration frontier. If clicking an element modifies the state of the \textit{current page} by expanding an interface, then we add the newly revealed elements as the clicked element's dropdown items.
After exploring the homepage, we repeat the above process for the pages that were added to the exploration frontier. For each page visited, we extract its title and summarize it. Exploration continues depth-first until a set maximum search depth is reached. We also limit the maximum number of elements explored per page, the total number of pages explored, and also set a timeout for each website.

\paragraph{Use at inference.}
$\mathcal{M}_w$ is saved per-site as JSON and reused across tasks. This memory is consumed by the agent in a highly token-efficient manner: rather than loading the full memory into the context window or retrieving large passages of text, only a handful of extra tokens are added per prompt. At task start the agent may select a small bookmark set $B_\tau \subseteq \mathcal{M}_w$ that remains available as navigation shortcuts throughout the task, and the agent's observation space is augmented to provide context about hidden dropdown menu elements.

We adopt a deliberately minimal memory instantiation in this work in order to efficiently demonstrate how our framework can be used to create structured site-specific memories for LLM agents. Our website representation could also be used as a building block to implement more advanced memory approaches such as those explored in Agent Workflow Memory \citep{wang2024agentworkflowmemory} or SkillWeaver \citep{zheng2025skillweaverwebagentsselfimprove}. We leave this to future work.

\subsection{Divide-and-Conquer Observation}
\label{sec:observation}

Large pages can easily exceed the reliable context window of small LLMs, and even within that window, flattening a full accessibility tree into a single prompt dilutes task-relevant signal among boilerplate. Our system addresses these issues by decomposing page analysis across multiple focused sub-prompts, extracting and condensing task-relevant information into a summarized observation $\hat{o}_t$.

\paragraph{PageMem retrieval and update.}
For the current page $p_t$, the agent first checks whether a PageMem already exists in $\mathcal{M}_{w}$ (either built during exploration or cached from a previous visit in the same task). If so, it is reused; otherwise a fresh PageMem is constructed from the live DOM as described in \S\ref{sec:pagemem}. When a cached PageMem is reused, sections whose elements have changed since the last visit are re-summarized while unchanged sections retain their cached summaries and extractions, amortizing summarization cost across timesteps and repeat visits.

\paragraph{Section selection.}
The LLM is shown the list of section summaries for $p_t$ along with the task instruction $I$ and the interaction history $h_t$, and returns a subset of sections $S_t \subseteq \{s_1, \ldots, s_n\}$ judged relevant to the task.

\paragraph{Detail extraction.}
For each PageSection $s \in S_t$, the LLM is prompted to extract task-relevant information from the section's full content (accessibility sub-tree, page metadata). If a section contains visible images above a minimum size, the URLs and VLM descriptions for the images are included in the extraction prompt. When $s$ is a table or list section, extraction is preceded by an item-selection step: items are grouped into chunks of maximum size $c$, the LLM selects relevant items from each chunk, and only the selected items are passed to the extraction prompt (Figure~\ref{fig:main}, middle-right). This chunked filtering keeps even very long lists or tables within context. Extractions are cached on the section and reused while the section is unchanged. Full process is in Appendix~\ref{app:observation}.

\paragraph{Summary synthesis.}
Finally, we provide the LLM with the extracted outputs from all selected sections and prompt it to generate a compact page summary $\hat{o}_t$, which becomes the observation passed to the action module and appended to the history $h_{t}$. We instruct the LLM to generate a one paragraph long summary, as we find this is sufficient in most cases to capture the task-relevant page information while also allowing the history representation to remain compact.

\subsection{Compound Actions and Workflows}
\label{sec:actions}

At each timestep $t$ the action module selects an action $a_t$ from a candidate set $\mathcal{A}_t$ assembled from the current PageMem, the selected sections $S_t$, and the agent's memory. Rather than exposing actions through an LLM tool-use interface, we present $\mathcal{A}_t$ as a numbered list and prompt the model to return the index of its chosen action, as we find this to be more reliable than tool use for small open-weight models. Our system then automatically executes the appropriate action function based on the action selected. Many actions are \emph{compound}: their execution invokes a workflow $\omega(a_t)$ that combines multiple LLM sub-calls with browser operations (implemented using Playwright\footnote{\url{https://playwright.dev/python/}}) to complete a multi-step interaction as a single action.

\paragraph{Action selection.}
$\mathcal{A}_t$ comprises three groups. \emph{Navigation} actions include previously visited URLs, bookmarks in $B_\tau$, a type-URL action, and (when applicable) switch-tab and switch-website. \emph{Element} actions are gathered from the selected sections $S_t$ and a rule-based pre-filter removes no-ops and rarely-useful actions (e.g., navigating to the current page, clicking a selected radio). The LLM filters the full list of navigation and element actions for those it deems most promising for the task. \emph{End task} is always available. The LLM then selects the next action $a_t$ from the filtered candidate set.

\paragraph{Workflows.}
Our design principle for workflows is that action sequences whose intermediate steps produce only \emph{partial} state changes to the current page (dropdown expansion, search suggestions, field-by-field form entry) are collapsed into a single compound action handled by a workflow, while actions that navigate to a different page are kept as individual agent actions. This keeps the agent's decision loop anchored at semantically meaningful transitions rather than at every micro-interaction. We describe two representative workflows; the full set is summarized in Appendix~\ref{app:workflows}.

\emph{Dropdown selection.} The workflow clicks the dropdown element, extracts the list of revealed options via a section-level diff, prompts the LLM to choose one option by index, and clicks the chosen option.

\emph{Form submission.} The LLM first selects which fields of the form to fill, is then prompted for each field's value (which is entered on the page), and finally reviews the completed form to either edit further or submit (Figure~\ref{fig:main}, bottom-right). The workflow handles field-specific details internally so that the agent-level action is a single \texttt{SubmitForm} step.

\paragraph{End-task.}
The end-task action invokes a one-time verification workflow: the LLM is prompted with the task and history and asked to either produce a final answer or report that the task is not yet complete, in which case the LLM is re-prompted to select a different non-terminating action.

\begin{table*}[t]
    \centering
    \small
    \caption{Benchmark success rates (\%).
WebChallenger sets new open-model SOTA on four web navigation benchmarks and performs comparably to agents built on proprietary models, despite using no training. Best proprietary and open-model results are bolded. VWA: VisualWebArena
\citep{koh2024visualwebarena}, O-M2W: Online-Mind2Web
\citep{xue2025onlinemind2web}, WoA: WorkArena
\citep{drouin2024workarenacapablewebagents}.}
    \label{tab:main-results}
    \begin{threeparttable}
    \setlength{\tabcolsep}{3pt}
    \begin{tabular}{lllcccccc}
        \toprule
        & \textbf{Method} & \textbf{Model(s)} & \textbf{WebArena} & \textbf{VWA} & \textbf{O-M2W} & \textbf{WoA L1} \\
        \midrule

        \multirow{3}{*}{\shortstack[l]{Proprietary \\ Models}}
         & GenericAgent & GPT-4o & 31.4 & 26.7 & - & 45.5 \\
         & GenericAgent & Claude 3.5 Sonnet & 36.2 & 21.0 & - & 56.4 \\
         & GenericAgent & GPT-5 & - & - & - & \textbf{79.1} \\
         & GenericAgent & GPT-4o-mini & 17.4 & 16.9 & - & 27.0 \\
         & WALT & GPT-5 & 50.1 & \textbf{52.9} & - & - \\
         & IBM CUGA & - & \textbf{61.7} & - & - & - \\
         & OpenAI CUA & - & 58.1 & - & \textbf{61.3} & - \\
         & ScribeAgent & GPT-4o + Qwen2.5-32B & 53.0 & - & - & - \\
         & AgentSymbiotic & Claude 3.5 + Llama3 8B & 48.5 & - & - & - \\
         & AgentOccam-Judge & GPT-4 & 45.7 & - & - & - \\
         & WebPilot & GPT-4o & 37.2 & - & - & - \\
         & Agent Workflow Memory & GPT-4o & 35.5 & - & - & - \\
         & SkillWeaver & GPT-4o & 29.8 & - & - & - \\

        \addlinespace[0.5em]
        \midrule

        \multirow{4}{*}{\shortstack[l]{Open-Source \\ Models \\ (fine-tuned)}}
         & Agent-as-Annotators & Qwen3.5-9B & 41.5 & 33.9 & - & 51.5 \\
         & Mobile-Agent-v3.5 & Qwen3-VL-32B & 48.4 & 46.6 & - & - \\
         & WebDreamer & Qwen2-VL-7B & - & 21.9 & 35.0 & - \\
         & Fara-7B & Qwen2.5-VL-7B & - & - & 34.1 & - \\
         & Learn-by-Interact & Codestral 22B & 24.2 & - & - & - \\
         & AgentTrek         & Qwen2.5-32B   & 22.4 & - & - & - \\
         & Go-Browse         & Qwen2.5-7B   & 21.7 & - & - & - \\
         & AutoWebGLM        & ChatGLM3 6B   & 18.2 & - & - & - \\
         & TTI & Gemma 3 12B & 26.1 & - & - & - \\

        \hdashline
        \noalign{\vskip 1.0ex}

        \multirow{2}{*}{(zero-shot)}
         & GenericAgent & GPT-oss-120b & - & - & - & 50.9 \\
         & Tree Search & Llama-3-70B-Instruct & 10.1 & 16.7 & - & - \\
         \addlinespace[0.5em]
         & \textit{\textbf{WebChallenger (Ours)}} & \shortstack[l]{GLM-4-32B + \\ Qwen2.5-VL-7B} & \textbf{56.3} & \textbf{48.7}\tnote{$\dagger$} & \textbf{51.0} & \textbf{70.9} \\

        \bottomrule
    \end{tabular}
    \begin{tablenotes}[flushleft]
        \footnotesize
        \item[$\dagger$] Our VisualWebArena experiments use Qwen3-VL-4B-Instruct in place of Qwen2.5-VL-7B-Instruct.
    \end{tablenotes}
    \end{threeparttable}
\end{table*}

\section{Experiments}
We evaluate WebChallenger on four open-ended web navigation benchmarks to test its performance on a diverse range of capabilities. \textbf{WebArena} \citep{zhou2024webarena} consists of 812 tasks in 6 simulated environments that are designed to mimic common website types (e.g., forum, wiki) and uses a combination of both programmatic and LLM evaluation. \textbf{VisualWebArena} \citep{koh2024visualwebarena} builds on the infrastructure of WebArena, but consists of 910 tasks that require visual reasoning. \textbf{Online-Mind2Web} \citep{xue2025onlinemind2web} consists of 300 tasks across 136 real-world websites. We score our agent using human evaluations for Online-Mind2Web. \textbf{WorkArena} \citep{drouin2024workarenacapablewebagents} contains 330 enterprise-related tasks that require agents to navigate complex user interfaces.

\subsection{Experimental Setup}
We use \texttt{GLM-4-32B-0414} \citep{glm4} as the LLM controller, and \texttt{Qwen2.5-VL-7B-Instruct} \citep{qwenvl} as our supplementary vision model for image captioning. For VisualWebArena, we use \texttt{Qwen3-VL-4B-Instruct} \citep{bai2025qwen3vltechnicalreport} as the vision model. For all experiments we use the same agent prompts and sample with temperature 0. For each benchmark, we first explore the full set of benchmark websites before running inference. During inference, the agent's memory is reset between tasks to the post-exploration state to preserve independence between evaluation samples. Additional experiment details are provided in Appendix~\ref{app:experiment-details}.

\subsection{Main Results}
\textbf{Baselines}. We compare WebChallenger against strong open model and proprietary baselines for each of our selected benchmarks.

For proprietary model baselines, we use WALT \citep{prabhu2025waltwebagentslearn}, IBM CUGA \citep{ibmcuga}, OpenAI CUA \citep{openaicua}, ScribeAgent \citep{shen2024scribeagentspecializedwebagents}, AgentSymbiotic, \citep{zhang2025symbioticcooperationwebagents}, AgentOccam-Judge \citep{yang2025agentoccamsimplestrongbaseline}, WebPilot \citep{zhang2024webpilotversatileautonomousmultiagent}, SkillWeaver \citep{zheng2025skillweaverwebagentsselfimprove}, and Agent Workflow Memory \citep{wang2024agentworkflowmemory}.

For open-model baselines, we use Agent-as-Annotators \citep{lu2026structureddistillationwebagent}, Mobile-Agent-v3.5 \citep{xu2026mobileagentv35multiplatformfundamentalgui}, WebDreamer \citep{gu2025llmsecretlyworldmodel}, Fara-7B \citep{awadallah2025fara7befficientagenticmodel}, Learn-by-Interact \citep{su2025learnbyinteractdatacentricframeworkselfadaptive}, AgentTrek \citep{xu2025agenttrekagenttrajectorysynthesis}, Go-Browse \citep{gandhi2025gobrowsetrainingwebagents}, AutoWebGLM \citep{lai2024autowebglmlargelanguagemodelbased}, TTI \citep{shen2025thinkingvsdoingagents}, and Tree Search \citep{koh2025treesearchlanguagemodel}.

GenericAgent results are taken from the official BrowserGym leaderboard \citep{browsergymleaderboard}. All other baseline results in Table \ref{tab:main-results} are taken from their original reports.

\textbf{Results}. As shown in Table~\ref{tab:main-results}, WebChallenger sets
new state-of-the-art results among open-model agents on all four benchmarks
despite using no fine-tuning. On WebArena, our 56.3\% exceeds the strongest
fine-tuned open-model baseline (Mobile-Agent-v3.5, 48.4\%) by 7.9 points and
surpasses ScribeAgent (53.0\%, GPT-4o planner). On VisualWebArena, 48.7\%
outperforms all open-model baselines and trails only WALT (52.9\%, GPT-5). On WorkArena,
70.9\% lands 20 points above the next-best zero-shot open model
and exceeds both Claude 3.5 Sonnet (56.4\%) and GPT-4o (45.5\%) backbones. 51.0\% on Online-Mind2Web shows that our framework generalizes by exploiting structural patterns shared across the web rather than site-specific adaptations.
These results demonstrate that careful architectural scaffolding
can close most of the gap between small open-weight models and frontier
proprietary systems on long-horizon web tasks, and that a single
configuration generalizes consistently across a wide range of tasks and environments.

\begin{table}  
    \centering
    \small
    \caption{Component ablations on WebArena-lite (165 tasks). Per-site task counts: Shopping ($n=46$), Reddit ($n=21$), GitLab ($n=32$), Maps ($n=31$), CMS ($n=35$). $\Delta$ is the change in average success rate relative to the full system.}
    \label{tab:component-ablations}
    \begin{tabular}{lcccccrr}
        \toprule
        \textbf{Method} & \textbf{Shopping} & \textbf{Reddit} & \textbf{GitLab} & \textbf{Maps} & \textbf{CMS} & \textbf{Avg} & \textbf{$\Delta$} \\
        \midrule
        \textit{WebChallenger} (GLM-4-32B) & 56.5 & 71.4 & 65.5 & 45.1 & 60.0 & \textbf{58.8} & -- \\
        \quad $-$ memory & 43.4 & 71.4 & 53.1 & 48.4 & 54.2 & 51.2 & $-7.6$ \\
        \quad $-$ compound actions & 47.8 & 66.7 & 50.0 & 48.4 & 40.0 & 49.1 & $-9.7$ \\
        \quad $-$ observation pipeline & 48.9 & 57.1 & 53.1 & 26.7 & 27.3 & 41.2 & $-17.6$ \\
        \bottomrule
    \end{tabular}
\end{table}

\subsection{Analysis}

We run additional experiments on the 165-task WebArena-lite subset \citep{liu2024visualagentbenchlargemultimodalmodels} to examine component contributions, compute usage, and backbone sensitivity; our system's lite score ($58.8$) tracks the full WebArena score ($56.3$) closely.

\paragraph{Component ablations.} We separately remove each of the three architectural components (Table~\ref{tab:component-ablations}). \textit{Remove memory} disables bookmarks, dropdown information, and pre-cached section summaries; PageMem is still used at inference time but constructed from scratch for each task. \textit{Remove compound actions} restricts the agent to single basic actions (\textsc{ClickElement}, \textsc{EnterInput}, \textsc{SelectOption}, \textsc{UploadFile}, plus navigation), eliminating the search, dropdown, and form-filling workflows. \textit{Remove observation pipeline} replaces section selection and detail extraction with a single prompt containing the full ax-tree and all available actions, with history reduced to a list of prior actions.

Among the three components, removing the observation pipeline causes the largest accuracy drop ($-17.6$ points), followed by compound actions ($-9.7$) and memory ($-7.6$). Compound action removal has its largest effect on CMS ($-20.0$), as CMS involves interactions with complex interfaces such as forms and filtering menus. On Reddit, removing memory has no effect ($71.4$ in both conditions), suggesting GLM-4-32B navigates Reddit reliably without pre-cached information. Maps performance is largely unaffected by memory and compound actions, as the Maps environment is focused on a single interface that doesn't benefit from those components.

\begin{table}  
    \centering
    \small
    \caption{Token and step usage for the GLM-4-32B component ablations. Tokens/Prompt is the average input token count per LLM call. We count compound actions as one step.}
    \label{tab:token-efficiency}
    \begin{tabular}{lrrr}
        \toprule
        \textbf{Method} & \textbf{Avg Steps} & \textbf{Total Tokens} & \textbf{Tokens/Prompt} \\
        \midrule
        \textit{WebChallenger} (GLM-4-32B) & 7.20 & 47.0M & 1850 \\
        \quad $-$ memory & 7.34 & 48.1M & 1781 \\
        \quad $-$ compound actions & 9.85 & 64.9M & 2109 \\
        \quad $-$ observation pipeline & 11.26 & 36.0M & 8793 \\
        \bottomrule
    \end{tabular}
\end{table}

\begin{table}  
    \centering
    \small
    \caption{Backbone model comparison on WebArena-lite \citep{liu2024visualagentbenchlargemultimodalmodels}. The bottom row uses the minimal  GenericAgent harness from BrowserGym \citep{dechezelles2025browsergymecosystemwebagent} with the same GLM-4-32B model used in our system to isolate the contribution of our harness.}
    \label{tab:backbone-comparison}
    \begin{tabular}{lcccccc}
        \toprule
        \textbf{Method} & \textbf{Shopping} & \textbf{Reddit} & \textbf{GitLab} & \textbf{Maps} & \textbf{CMS} & \textbf{Avg} \\
        \midrule
        \textit{WebChallenger} w/ GPT-5 & 60.9 & 71.4 & 70.8 & 74.2 & 70.6 & \textbf{68.7} \\
        \textit{WebChallenger} w/ GLM-4-32B & 56.5 & 71.4 & 65.5 & 45.1 & 60.0 & 58.8 \\
        \textit{WebChallenger} w/ GPT-4o-mini & 36.9 & 57.1 & 43.7 & 51.6 & 51.4 & 46.7 \\
        \midrule
        GenericAgent w/ GLM-4-32B & 19.5 & 23.8 & 28.1 & 16.1 & 11.4 & 19.4 \\
        \bottomrule
    \end{tabular}
\end{table}

\paragraph{Token and step efficiency.} Removing the observation pipeline reduces total tokens ($47.0$M $\rightarrow$ $36.0$M) but raises average prompt size $4.75\times$ ($1850 \rightarrow 8793$ tokens) and step count from $7.2$ to $11.26$ (Table~\ref{tab:token-efficiency}). Our multi-stage observation processing decomposes large difficult prompts into several smaller but easier prompts, trading inference compute for performance. Compound actions significantly improve agent efficiency: removal causes total tokens to rise to $64.9$M and steps to $9.85$, since interactions that previously executed within a single workflow now require a separate observation and decision cycle per atomic action.

\paragraph{Backbone model comparison.} We swap the GLM-4-32B backbone for GPT-5 and GPT-4o-mini, and additionally evaluate GLM-4-32B alone in the minimal GenericAgent harness to isolate the architecture's contribution (Table~\ref{tab:backbone-comparison}). GPT-5 reaches $68.7\%$, $9.9$ points above GLM-4-32B in the same framework. GPT-4o-mini reaches $46.7\%$, indicating our framework retains strong performance even with weaker backbones. GLM-4-32B in the GenericAgent harness scores $19.4\%$, against $58.8\%$ in our framework, a $39.4$-point improvement from system architecture alone.

\section{Related Work}
\textbf{Agent Memory}. A growing body of work equips LLM web agents with external memory by accumulating insights from task trajectories \citep{wang2024agentworkflowmemory, ouyang2025reasoningbankscalingagentselfevolving, sarch2025vlmagentsgeneratememories, pang-etal-2025-assimilation, liu2025webcoachselfevolvingwebagents, nekoei2025justintimeepisodicfeedbackhinter, fu2024autoguideautomatedgenerationselection, chen2024automanualconstructinginstructionmanuals, cheng2025webatlasllmagentexperiencedriven, su2025learnbyinteractdatacentricframeworkselfadaptive}. Our memory takes a complementary route: a deterministic exploration procedure efficiently produces a structured site map with no task experience, demonstrations, or documentation required, making it applicable to any website out of the box.

\textbf{Web Action Space}. Several works extend web agent action spaces beyond click and type by introducing higher-level programmatic skills \citep{song2025browsingapibasedwebagents, wang2025inducingprogrammaticskillsagentic, zheng2025skillweaverwebagentsselfimprove, he2025reconactselfevolvingmultiagentbrowseruse, prabhu2025waltwebagentslearn, yu2025polyskilllearninggeneralizableskills, wang2026webxskillskilllearningautonomous, zhong2026actionenginereactiveprogrammaticgui}. These approaches typically learn site-specific code, whereas our compound workflows operate over PageMem's abstract elements and sections and generalize across sites with no per-site adaptation. We also depart from the standard tool-calling interface in favor of a numbered-list action format.

\textbf{Observation Refinement}. Web agents observe their environment through text \citep{gur2018learningnavigateweb, li2023zeroshotlanguageagentcomputer, kim2023languagemodelssolvecomputer}, screenshots \citep{shaw2023pixelsuiactionslearning, hong2024cogagentvisuallanguagemodel, gou2025navigatingdigitalworldhumans, pahuja2025explorerscalingexplorationdrivenweb, he2024webvoyagerbuildingendtoendweb, zheng2024gpt4visiongeneralistwebagent, verma2024adaptagentadaptingmultimodalweb}, or both \citep{furuta2024multimodalwebnavigationinstructionfinetuned}. All such modalities are token-heavy and information-sparse, motivating refinement strategies. Text-based agents prune irrelevant HTML elements \citep{gur2024realworldwebagentplanninglong, deng2023mind2webgeneralistagentweb, kil2024dualviewvisualcontextualizationweb, lù2024weblinxrealworldwebsitenavigation, lee2025learningcontextualizewebpages, abuelsaad2024agenteautonomouswebnavigation, kerboua2025focusagentsimpleeffectiveways}, while vision-based agents focus attention on specific screen regions \citep{sarch2025groundedreinforcementlearningvisual, singh2025trishulregionidentificationscreen, luo2025visualtesttimescalinggui, park2025rvlmregionawarevisionlanguage}. We apply region-based focus to a hybrid text-vision agent by splitting pages along DOM structure, which preserves the semantic grouping authored into the page better than pixel-space cropping. \citet{feuillademontixi2026webfurl} explores a similar DOM-based approach. More broadly, our pipeline echoes a line of work on decomposing long-context tasks into focused sub-prompts \citep{zhang2026recursivelanguagemodels, chen2023walkingmemorymazecontext, jayalath2025prismefficientlongrangereasoning, lee2024humaninspiredreadingagentgist}.

\section{Conclusion}
WebChallenger closes much of the gap between small open-weight models and
frontier proprietary systems on long-horizon web navigation. We argue current LLMs already possess sufficient intelligence
for many common web tasks, but standard frameworks fail to scaffold that intelligence
with the selective attention, persistent memory, and procedural fluency
humans rely on. We supply each through a divide-and-conquer observation
pipeline, an offline exploration and memory system, and compound action
workflows. These components are implemented on top of PageMem, a shared page representation that
generalizes across websites without site-specific adapters. Using small, general-purpose models without fine-tuning, our system sets new state-of-the-art results among open-weight agents on four diverse web agent benchmarks.

\section*{Acknowledgements}
We thank the ML Collective community for their support, discussions, and feedback.


\bibliography{references}
\bibliographystyle{icml2025}


\appendix

\section{Implementation Details}

\subsection{PageMem}
\label{app:pagemem}
We provide the details of our memory structure and describe the PageMem construction process. Each PageMem is built in two stages: \textsc{DividePage} (Algorithm~\ref{alg:dividepage}) recursively partitions the live DOM tree into an ordered list of empty PageSections forming the structural skeleton of the page, and \textsc{UpdatePageMem} (Algorithm~\ref{alg:updatepagemem}) then populates each section with its interactable elements and an LLM-generated summary, and finally generates the page-level summary.

\subsubsection{Memory Structure}

We provide a detailed formalization of the memory hierarchy from \S\ref{sec:pagemem}. Throughout, we use $\sigma$ for model-generated summaries, $\alpha$ for immutable DOM-derived attributes, and $\mu$ for mutable agent-side state accumulated during exploration and task execution. Descriptions of $\alpha$ and $\mu$ below give representative fields rather than exhaustive listings; full field schemas are documented in our released code.

\paragraph{WebsiteMem.}
A WebsiteMem for website $w$ is a tuple
\[
\mathcal{M}_w = (P_w,\, T_w,\, E_w),
\]
where $P_w$ is a mapping from URL to PageMem, collecting all concrete pages encountered on $w$; $T_w$ is a list of list-page templates, each itself a PageMem, against which newly visited pages are matched by structural comparison (see~\ref{template-match}); and $E_w$ is the set of all elements encountered on $w$, used to deduplicate elements during exploration.

\paragraph{PageMem.}
A PageMem is a tuple
\[
p = (u_p,\, n_p,\, \sigma_p,\, S_p,\, \mu_p),
\]
where $u_p$ is the page URL; $n_p$ is the page title; $\sigma_p$ is a VLM-generated page-level summary; $S_p = (s_1, \ldots, s_{|S_{p}|})$ is an ordered list of PageSections; and $\mu_p$ holds page-level agent state (e.g., information extracted by agent, the agent's past interaction history on the page).

\paragraph{PageSection.}
A PageSection is a tuple
\[
s = (\sigma_s,\, E_s,\, S'_s,\, \alpha_s,\, \mu_s),
\]
where $\sigma_s$ is a VLM-generated section summary; $E_s = (e_1, \ldots, e_{k_s})$ is an ordered list of Elements contained in the section; $S'_s = (s'_1, \ldots, s'_{m_s})$ is an ordered list of sub-sections, empty for normal sections and containing one sub-section per item for list sections; $\alpha_s$ holds DOM-derived attributes used for creating selectors (e.g., id, tag, class, DOM-subtree handle, bounding box) and $\mu_s$ holds mutable agent state (e.g., task-relevant extractions, VLM-generated image descriptions, a staleness flag indicating whether the DOM subtree has changed since $\sigma_s$ was last computed).

\paragraph{Element.}
An Element is a tuple
\[
e = (\alpha_e,\, E'_e,\, \mu_e),
\]
where $\alpha_e$ holds DOM-derived attributes (e.g., id, tag, class, role, label, type); $E'_e = (e'_1, \ldots, e'_{l_e})$ is an ordered list of dropdown items (themselves Elements), which is empty for non-dropdown elements; and $\mu_e$ holds mutable agent state (e.g., the element's current input value, a flag for whether the agent has clicked the element during the current task).

\subsubsection{Page division.}
We provide the pseudocode for page splitting in Algorithm \ref{alg:dividepage}. \textsc{DividePage} takes the root of the DOM tree and returns a PageMem whose ordered section list forms the structural skeleton of the page. The procedure recursively descends the DOM, terminating at nodes that form a meaningful grouping, either semantically (by tag), visually (by size), or structurally (by repetition of siblings). Groups of $\geq 4$ consecutive siblings sharing tag and class are merged into a single list section node before recursion; list sections are always terminal and are never further subdivided.

\begin{algorithm}[h]
\caption{\textsc{DividePage}}
\label{alg:dividepage}
\begin{algorithmic}[1]
\Require DOM root node $r$
\Ensure PageMem $p$
\State $L \gets [\,]$
\State \Call{Split}{$r$, $L$}
\State $p \gets \Call{NewPageMem}{\,}$
\State $p.u \gets \Call{CurrentURL}{\,}$
\State $p.n \gets \Call{ExtractTitle}{\,}$
\State $p.S \gets L$
\State \Return $p$
\Statex
\Procedure{Split}{node $v$, list $L$}
  \If{\Call{IsTerminal}{$v$}}
    \State append \Call{MakeSection}{$v$} to $L$
  \Else
    \State $C \gets \Call{GroupSiblings}{v.\text{children}}$
    \For{$c \in C$}
      \State \Call{Split}{$c$, $L$}
    \EndFor
  \EndIf
\EndProcedure
\Statex
\Function{IsTerminal}{node $v$}
  \State \Return $v.\text{isListSection} \ \lor\ v.\text{tag} \in \mathcal{T}_{\text{group}} \ \lor\ \lnot\,\Call{Oversized}{v}$
\EndFunction
\Statex
\Function{Oversized}{node $v$}
  \State \Return $(v.h > 900 \land v.w > 320) \ \lor\ (v.h > 500 \land v.w > 800)$
\EndFunction
\Statex
\Function{GroupSiblings}{children $(c_1, \ldots, c_k)$}
  \State scan $(c_1, \ldots, c_k)$ for groups of consecutive siblings that share tag and class
  \State replace each group of length $\geq 4$ with a single list-section node containing the group
  \State \Return the resulting (shortened) sequence
\EndFunction
\end{algorithmic}
\end{algorithm}

\paragraph{Parameters.}
The grouping tag set is
\[
\begin{split}
\mathcal{T}_{\text{group}} = \{ & \texttt{ol}, \texttt{ul}, \texttt{table}, \texttt{form}, \texttt{fieldset}, \texttt{aside}, \texttt{article}, \\
& \texttt{details}, \texttt{p}, \texttt{img}, \texttt{embed}, \texttt{code}, \texttt{group}, \texttt{nav}, \texttt{header}, \texttt{footer} \}.
\end{split}
\]
Dimensions $v.h$ and $v.w$ in \textsc{Oversized} are the node's rendered bounding-box height and width in CSS pixels, obtained from the browser's layout engine.

\subsubsection{PageMem update.}
\textsc{UpdatePageMem} refreshes a PageMem to reflect the current live page state and is invoked at the start of every observation step. It is also the routine that populates a freshly divided PageMem with its initial elements and summaries.

\textsc{UpdateSection} (i) queries the browser for the section's current set of interactable elements, (ii) computes the added / removed / modified diff $\Delta = (\Delta^+, \Delta^-, \Delta^\sim)$ against the section's previous element list, (iii) re-summarizes if the section has no summary yet or if the structural change is large enough, and (iv) returns $\Delta$ so that the caller (an observation step or a workflow) can respond to partial state changes. This is the machinery behind, e.g., the dropdown workflow of \S\ref{sec:actions}, which reads the revealed options directly from $\Delta^+$. List sections are handled specially: their content is unbounded and repetitive, so instead of enumerating elements or tracking a diff at the list level, we always re-summarize from a screenshot and return an empty diff. Element-level tracking happens only on the per-item sub-sections, and only after list-item selection in the observation pipeline (\S\ref{sec:observation}).

\begin{algorithm}[h]
\caption{\textsc{UpdatePageMem} and \textsc{UpdateSection}}
\label{alg:updatepagemem}
\begin{algorithmic}[1]
\Procedure{UpdatePageMem}{PageMem $p$}
  \For{$s \in p.S$}
    \State \Call{UpdateSection}{$s$}
  \EndFor
  \If{$p.\sigma_p$ is undefined}
    \State $p.\sigma_p \gets \Call{VLMSummarizePage}{p}$
  \EndIf
\EndProcedure
\Statex
\Procedure{UpdateSection}{PageSection $s$}
  \If{$s$ is a list section} \Comment{list sections only get summarized}
    \State $s.\sigma_s \gets \Call{VLMSummarizeSection}{s}$
    \State \Return $(\emptyset, \emptyset, \emptyset)$
  \EndIf
  \State $E_{\text{new}} \gets \Call{GetElements}{s}$
  \State $\Delta \gets \Call{Diff}{s.E_s,\, E_{\text{new}}}$ \Comment{$\Delta^+$: added, $\Delta^-$: removed, $\Delta^\sim$: input-value changes}
  \State $s.E_s \gets E_{\text{new}}$
  \If{$s.\sigma_s$ is undefined}
    \State $s.\sigma_s \gets \Call{VLMSummarizeSection}{s}$
  \ElsIf{$|\Delta^+| + |\Delta^-| \geq 3$} \Comment{re-summarize if $\geq 3$ elements added/removed}
    \State $s.\sigma_s \gets \Call{VLMSummarizeSection}{s}$
  \EndIf
  \State \Return $\Delta$
\EndProcedure
\end{algorithmic}
\end{algorithm}

\paragraph{Element population.}
\textsc{GetElements} produces the element list for a section via a three-step pipeline. A helper first resolves the section to a Playwright locator. The locator's descendants are then filtered by the clickable predicate \textsc{IsClickable} defined below. Finally, each surviving DOM node is passed to an Element constructor that reads its DOM attributes into $\alpha_e$.

\paragraph{Clickable predicate.}
A DOM node $v$ is considered interactable iff it passes a visibility-and-accessibility gate \emph{and} satisfies at least one positive signal. The gate excludes nodes that are not rendered, carry the \texttt{disabled} attribute, or have \texttt{aria-hidden="true"}. The positive signals are any of: a tag in an interactable tag set, a DOM event-listener attribute in a listener set, an ARIA role in an interactable role set, or a computed \texttt{cursor} style of \texttt{pointer}. Formally,
\[
\begin{split}
\textsc{IsClickable}(v) \ \equiv\ \big(v.\text{tag} \in \mathcal{T}_{\text{clk}}\ \lor\ v.\text{attrs} \cap \mathcal{L}_{\text{clk}} \neq \emptyset\ \lor\ v.\text{role} \in \mathcal{R}_{\text{clk}}\ \lor\ v.\text{cursor} = \texttt{pointer}\big) \\
\land\ \textsc{Accessible}(v),
\end{split}
\]
with the sets
\begin{align*}
\mathcal{T}_{\text{clk}} &= \{\texttt{button}, \texttt{a}, \texttt{input}, \texttt{select}, \texttt{textarea}, \texttt{details}, \texttt{summary}, \texttt{option}\}, \\
\mathcal{L}_{\text{clk}} &= \{\texttt{onclick}, \texttt{onmousedown}, \texttt{onmouseup}, \texttt{onkeydown}, \texttt{onkeyup}\}, \\
\mathcal{R}_{\text{clk}} &= \{\texttt{button}, \texttt{link}, \texttt{menuitem}, \texttt{option}, \texttt{radio}, \texttt{checkbox}, \texttt{tab}, \\
                           &\qquad \texttt{textbox}, \texttt{combobox}, \texttt{slider}, \texttt{spinbutton}, \texttt{search}, \texttt{searchbox}\}.
\end{align*}
These heuristics are adapted from BrowserUse~\citep{browser-use}.

\subsection{Exploration}
\label{app:exploration}

We provide the details of the offline exploration procedure that builds the WebsiteMem $\mathcal{M}_w$ used at inference. Exploration is a deterministic depth-first traversal of a website's pages and clickable elements, deduplicated against the running set $E_w$ of all elements seen on the site, with state restored between element clicks by reloading the pre-click URL. \textsc{ExplorePage} (Algorithm~\ref{alg:explorepage}) is the recursive driver that visits one page at a time; it delegates to \textsc{IteratePage} and \textsc{ExploreElement} (Algorithm~\ref{alg:iteratepage}) for the element-level work. Exploration is launched per website by initializing an empty $\mathcal{M}_w$ and a URL-only PageMem stub at a chosen starting URL --- the homepage in our experiments --- and invoking \textsc{ExplorePage} at the configured maximum depth. We abstract over the per-page element budget, total page budget, and per-website timeout in the pseudocode for clarity; these limits act as additional early-return checks throughout, and their values are reported per benchmark in Appendix~\ref{app:experiment-details}.

\begin{algorithm}[h]
\caption{\textsc{ExplorePage}}
\label{alg:explorepage}
\begin{algorithmic}[1]
\Procedure{ExplorePage}{PageMem stub $p$, depth $d$}
  \If{$p.u \in \mathcal{M}_w.P_w$} \Comment{URL already explored on this site}
    \State \Return
  \EndIf
  \State \Call{Navigate}{$p.u$}
  \State Populate $p$ with sections, elements and summaries
  \State add $p$ to $\mathcal{M}_w.P_w$ keyed by $p.u$
  \If{\Call{MatchesTemplate}{$p,\, \mathcal{M}_w.T_w$}} \Comment{already explored page with same structure}
    \State \Return
  \ElsIf{\Call{HasListSection}{$p$} $\lor\ p.\text{is\_list\_item}$}
    \State add $p$ to $\mathcal{M}_w.T_w$
  \EndIf
  \State $N \gets$ \Call{IteratePage}{$p$} \Comment{explore elements on page}
  \If{$d = 0$ \textbf{or} budget exhausted} \Return \EndIf
  \For{$p' \in N$}
    \State \Call{ExplorePage}{$p',\, d - 1$}
  \EndFor
\EndProcedure
\end{algorithmic}
\end{algorithm}

\paragraph{Page-level traversal.}
\textsc{ExplorePage} takes a stub PageMem carrying a target URL and the remaining recursion depth. It deduplicates against URLs already in $\mathcal{M}_w$, navigates the browser to the page, runs \textsc{DividePage} on the freshly-loaded DOM to construct the full PageMem, and \textsc{UpdatePageMem} to populate elements and summaries. The newly-built PageMem is then registered in $\mathcal{M}_w$. If its section structure matches an existing template in $T_w$, the page is treated as a known-shape duplicate and not iterated, since further iteration would re-cover element behaviors already learned from the matching template; otherwise, if the page contains a list section or its \texttt{is\_list\_item} flag is set, the PageMem is added to $T_w$ as a new template. \textsc{IteratePage} is then called on the page, returning a list of stub PageMems for newly-discovered URLs, which the procedure recursively explores at depth $d - 1$.

\paragraph{Template matching.}
\label{template-match}
\textsc{MatchesTemplate} compares the candidate PageMem against each template in $T_w$. Two PageMems match when they have the same number of sections and each pair of corresponding sections is structurally equivalent under the DOM-derived attributes in $\alpha_s$ (tag, class, and other selector-defining attributes). Because \textsc{DividePage} is deterministic on the DOM, structurally equivalent pages reliably yield identical section sequences in practice, so exact structural equality is sufficient as a match criterion without needing a similarity threshold. Matching is checked only against $T_w$ rather than all of $P_w$, both for efficiency and because non-template pages are by definition idiosyncratic and not expected to recur.

\paragraph{Element-level traversal.}
\textsc{IteratePage} walks the page's elements in document order. For elements not contained in any list section, it skips those already in the global element set $E_w$ and registers each new element in $E_w$. List sections are handled separately: rather than iterating every list item (which would redundantly re-cover elements with structurally identical neighbors), the procedure invokes \textsc{IterateListItem}, which iterates the elements contained in a single list-item container using the same per-element logic. Stubs returned from list-item exploration are tagged with \texttt{is\_list\_item} so that the recursive \textsc{ExplorePage} call can promote the resulting pages to templates.

\textsc{ExploreElement} returns the newly-discovered URL stub(s) reached by clicking the element. It first applies a static skip filter (described below) that rules out elements unsafe or unhelpful to click. It then records the pre-click URL, clicks the element, and inspects the result. If the URL has not changed, the post-click diff in page state is computed. Any newly-revealed elements ($\Delta^+$) are recorded as the clicked element's \texttt{dropdown\_elements} and recursively explored using the same \textsc{ExploreElement} routine. If the URL has changed and points to a same-site page not yet in $\mathcal{M}_w$, a URL-only stub is created and returned. Finally, the browser is reloaded to the pre-click URL to restore page state for the next iteration.

\begin{algorithm}[h]
\caption{\textsc{IteratePage} and \textsc{ExploreElement}}
\label{alg:iteratepage}
\begin{algorithmic}[1]
\Procedure{IteratePage}{PageMem $p$}
  \State $N \gets [\,]$
  \For{element $e$ in $p$ not contained in a list section}
    \If{per-page element budget exhausted} \textbf{break} \EndIf
    \If{$e \in \mathcal{M}_w.E_w$} \textbf{continue} \EndIf
    \State add $e$ to $\mathcal{M}_w.E_w$
    \State $N \gets N\, +$ \Call{ExploreElement}{$e$}
  \EndFor
  \For{list section $s \in p.S$}
    \State $N_\ell \gets$ \Call{IterateListItem}{$s$} \Comment{explore elements in one list-item container}
    \For{$p' \in N_\ell$} $p'.\text{is\_list\_item} \gets \text{true}$ \EndFor
    \State $N \gets N\, +\, N_\ell$
  \EndFor
  \State \Return $N$
\EndProcedure
\Statex
\Procedure{ExploreElement}{Element $e$}
  \State $N \gets [\,]$
  \If{\Call{ShouldSkip}{$e$}} \Return $N$ \EndIf
  \State $u_{\text{pre}} \gets$ \Call{CurrentURL}{\,}
  \State \Call{Click}{$e$}
  \State $u_{\text{post}} \gets$ \Call{CurrentURL}{\,}
  \If{$u_{\text{post}} = u_{\text{pre}}$}
    \State Identify newly revealed elements $\Delta^+$
    \If{$\Delta^+ \neq \emptyset$} \Comment{click revealed new elements (dropdown opened)}
      \State $e.\text{dropdown\_elements} \gets \Delta^+$
      \For{$e' \in \Delta^+$}
        \State $N \gets N\, +$ \Call{ExploreElement}{$e'$} \Comment{explore dropdown elements}
      \EndFor
    \EndIf
  \ElsIf{$u_{\text{post}}$ is on the same site $w$ \textbf{and} $u_{\text{post}} \notin \mathcal{M}_w.P_w$}
    \State create stub $p_{\text{new}}$ with $p_{\text{new}}.u \gets u_{\text{post}}$
    \State $N \gets N\, +\, [p_{\text{new}}]$
  \EndIf
  \State \Call{Navigate}{$u_{\text{pre}}$} \Comment{restore pre-click state for the next iteration}
  \State \Return $N$
\EndProcedure
\end{algorithmic}
\end{algorithm}

\paragraph{Skip filter.}
\textsc{ShouldSkip} excludes four categories of elements before any click is issued: (i) off-site links, identified by an \texttt{href} pointing to a domain outside $w$; (ii) authentication links such as login and sign-up, identified by keyword matching against the link text and URL path; (iii) \texttt{tel:}, \texttt{mailto:}, and \texttt{javascript:print(\dots)} links, identified by the \texttt{href} scheme; and (iv) \emph{modifier} buttons that could mutate persistent site state, identified either by the form attribute \texttt{type="submit"} or by keyword matching of the element's accessible text against destructive terms (\texttt{delete}, \texttt{remove}, \texttt{submit}, \texttt{save}, etc.).

\subsection{Observation Pipeline}
\label{app:observation}
 
We provide the details of the detail-extraction and summarization stages of the observation pipeline (\S\ref{sec:observation}). \textsc{AnalyzePage} (Algorithm~\ref{alg:analyzepage}) acts as the main driver: given the set of sections $S_t$ selected as relevant, it extracts task-relevant information from each and synthesizes a page summary. List sections are routed through \textsc{SelectListItems} (Algorithm~\ref{alg:selectlistitems}), which uses chunked LLM selection with explicit early termination to keep arbitrarily long lists within context.
 
\paragraph{Per-section detail extraction.}
For each selected section, a helper \textsc{Format} produces the \emph{details string} consumed by the extraction LLM. For a normal section, the details string contains the section's accessibility subtree together with the URLs and VLM-generated descriptions of any images in the section above the minimum size threshold (50 x 50 pixels). For a list section, \textsc{SelectListItems} is invoked first to choose a subset of items, and the details string is the per-item content formatted as a numbered list, with each entry containing the same accessibility-subtree-plus-image content as a normal section. The extraction call \textsc{LLMExtractDetails} (Prompt~\ref{prompt:ExtractDetails}) caches its output on the PageSection together with the details string used to produce it; on a subsequent call with an identical details string, the cached extraction is returned without an LLM call.
 
\begin{algorithm}[h]
\caption{\textsc{AnalyzePage}}
\label{alg:analyzepage}
\begin{algorithmic}[1]
\Procedure{AnalyzePage}{PageMem $p$, selected sections $S_t$}
  \State $X \gets [\,]$ \Comment{per-section extraction strings, in selection order}
  \For{$s \in S_t$}
    \If{$s$ is a list section}
      \State \Call{SelectListItems}{$s$} \Comment{populates $s$'s sub-sections and $s.E_s$}
    \EndIf
    \State $D \gets$ \Call{Format}{$s$}
    \State $x \gets$ \Call{LLMExtractDetails}{$D$} \Comment{returns cached value if $D$ unchanged}
    \State append \texttt{"<idx> <tag> <class>: "} $+\ x$ to $X$
  \EndFor
  \State $p.\text{task\_summary} \gets$ \Call{LLMSummarizePage}{$X$} \Comment{regenerated every call}
  \State \Return $p.\text{task\_summary}$
\EndProcedure
\end{algorithmic}
\end{algorithm}
 
\paragraph{List item selection.}
A list section can contain hundreds or thousands of items, far exceeding what fits in a single LLM context. \textsc{SelectListItems} addresses this by chunking the items sequentially into fixed-size groups and prompting the LLM to select relevant items chunk by chunk (Prompt~\ref{prompt:SelectItems}). After each chunk, a separate LLM call is issued (Prompt~\ref{prompt:CheckDone}) that sees the indices already searched, the items already selected, and the remaining entries, and decides whether to terminate early --- this avoids paying the cost of scanning the full list when the relevant items have already been found (e.g., the top few results of a sorted list). After selection, the procedure rebuilds the list section's sub-sections from the selected items, populating each via \textsc{UpdateSection}, and overwrites the list section's element list $E_s$ with only the elements from selected items. The original full element set is not retained: re-selection on a later observation step rebuilds the sub-sections from scratch from the live page state.
 
\begin{algorithm}[h]
\caption{\textsc{SelectListItems}}
\label{alg:selectlistitems}
\begin{algorithmic}[1]
\Procedure{SelectListItems}{list section $s$}
  \State $I^\star \gets [\,]$ \Comment{indices of items selected so far}
  \State partition the items of $s$ into sequential chunks $(C_1, C_2, \ldots)$ of fixed size $c$
  \For{$k = 1, 2, \ldots$}
    \State $I^\star \gets I^\star\, +$ \Call{LLMSelectItems}{$C_k,\, I^\star$}
    \If{\Call{LLMCheckDone}{$I^\star$, indices searched so far, remaining items}}
      \State \textbf{break}
    \EndIf
  \EndFor
  \State rebuild $s.S'_s$ as one PageSection per item index in $I^\star$
  \For{$s' \in s.S'_s$} \Call{UpdateSection}{$s'$} \EndFor
  \State $s.E_s \gets$ concatenation of $s'.E_{s'}$ over $s' \in s.S'_s$ \Comment{only selected items contribute actions}
\EndProcedure
\end{algorithmic}
\end{algorithm}
 
\paragraph{Summary caching.}
Two caches operate at different lifetimes within the observation pipeline. Section summaries $\sigma_s$ are populated by \textsc{UpdateSection} (Algorithm~\ref{alg:updatepagemem}) and persist across tasks within a WebsiteMem. Per-section task extractions $x$ produced by \textsc{LLMExtractDetails} are cached on the PageSection alongside the details string $D$ that produced them, and are reused for the lifetime of a task whenever the section's content is unchanged. The page-level task summary $p.\text{task\_summary}$ is always regenerated on each call to \textsc{AnalyzePage}, since the relevant framing of a page can shift as the task progresses through its history $h_t$.

\subsection{Agent Loop}
\label{app:agent-loop}
 
We provide the details of the top-level inference loop that integrates the observation pipeline (\S\ref{sec:observation}, App.~\ref{app:observation}) with the action system (\S\ref{sec:actions}). \textsc{AgentLoop} (Algorithm~\ref{alg:agentloop}) executes one timestep at a time until either the task is verified complete or the step budget is exhausted. Each timestep produces \emph{one} agent action, which may itself be a compound workflow that internally issues multiple LLM sub-calls and browser operations. After a non-navigating action, an intra-step continuation loop allows the agent to chain follow-up actions on the same page without re-running the full observation pipeline, up to a small budget. Bookmark and (where applicable) website pre-selection happen once at task start (Prompt~\ref{prompt:SelectBookmarks}).
 
\begin{algorithm}[h]
\caption{\textsc{AgentLoop}}
\label{alg:agentloop}
\begin{algorithmic}[1]
\Procedure{AgentLoop}{task $\tau = (I, u_0)$, WebsiteMem $\mathcal{M}_w$}
  \State $B_\tau \gets$ \Call{LLMSelectBookmarks}{$\mathcal{M}_w,\, I$} \Comment{optional, at task start only}
  \State $h \gets [\,]$
  \For{$t = 1, \ldots, T_{\max}$}
    \State $p \gets$ \Call{GetPageMem}{current URL, $\mathcal{M}_w$}
    \If{$m \gets$ \Call{CheckModal}{$p$}}
      \State $S \gets [m]$ \Comment{modal: focus on dialog, skip section selection}
    \Else
      \State $S \gets$ \Call{LLMSelectSections}{$p$}
    \EndIf
    \State $\hat{o} \gets$ \Call{AnalyzePage}{$p,\, S$}
    \State $u_{\text{pre}} \gets$ current URL
    \For{$j = 1, \ldots, J_{\max}$} \Comment{intra-step continuation, $J_{\max}=5$}
      \State $\mathcal{A} \gets$ \Call{GatherCandidates}{$p,\, S,\, p.S \setminus S,\, B_\tau$}
      \State $(a, r) \gets$ \Call{LLMSelectAction}{$\mathcal{A}$} \Comment{up to 3 retries on action error}
      \If{$a$ is end-task}
        \If{\Call{LLMVerifyEndTask}{$I,\, h$}} \Comment{1 verification check per task}
          \State \Return \Call{LLMFinalAnswer}{$I,\, h$}
        \Else
          \State remove end-task from $\mathcal{A}$ and re-prompt for $a$
        \EndIf
      \EndIf
      \State \Call{ExecuteAction}{$a$} \Comment{single op or compound workflow}
      \If{current URL $\neq u_{\text{pre}}$} \textbf{break} \EndIf
      \If{\Call{CheckModal}{$p$} $\neq \mathbf{nil}$} \textbf{break} \EndIf
      \State \Call{UpdatePageMem}{$p$}
      \State $\hat{o}' \gets \hat{o}\, +$ \Call{VLMScreenDiff}{$p$} \Comment{cheap update for follow-up action}
    \EndFor
    \State append step observation, reason, and action to $h$
  \EndFor
  \State \Return \Call{LLMFinalAnswer}{$I,\, h$} \Comment{step budget exhausted}
\EndProcedure
\end{algorithmic}
\end{algorithm}
 
\paragraph{Observation phase.}
At the start of each timestep, the agent retrieves or constructs the PageMem $p_t$ for the current page and refreshes it via \textsc{UpdatePageMem}. A modal-detection helper \textsc{CheckModal} then tests for the presence of a modal dialog using DOM heuristics including \texttt{role="dialog"} and \texttt{aria-modal="true"}; when a modal is detected, section selection is bypassed entirely and the modal's PageSection is used as the sole relevant section, focusing the agent's attention on the dialog and preventing the surrounding (now-inert) page from polluting the candidate space. Otherwise, the LLM selects relevant sections $S_t$ from the section summaries as in \S\ref{sec:observation} (Prompt~\ref{prompt:SelectSections}). The remaining sections $S_t^{\complement} = p_t.S \setminus S_t$ are kept aside for use during candidate assembly. \textsc{AnalyzePage} then produces the task summary $\hat{o}_t$.
 
\paragraph{Action phase.}
\textsc{GatherCandidates} assembles the candidate set $\mathcal{A}_t$ in two passes. First, for each $s \in S_t$, an LLM call selects elements from $s$ likely to be useful for the task, conditioned on $s$'s extracted details (Prompt~\ref{prompt:ElementCandidates}). Second, a single LLM call covers the elements of the first five entries of $S_t^{\complement}$ in document order --- a heuristic that ensures upper-page UI (navigation bars, search boxes, primary buttons) remains reachable even when the LLM did not flag those sections as relevant during section selection. Navigation actions (visited URLs, bookmarks $B_\tau$, type-URL, switch-tab, switch-website) are filtered by an LLM pass against the task (Prompt~\ref{prompt:NavigationCandidates}). The end-task action is always appended. A rule-based pre-filter removes irrelevant actions (e.g., switching to an already active tab, clicking an already-selected radio button, print, tel links, links leading outside allowed domains) before $\mathcal{A}_t$ is presented to the LLM.
 
The action-selection LLM call (Prompt~\ref{prompt:SelectAction}) returns a chosen action $a$ together with a natural-language \emph{reason} $r$ explaining the choice. \textsc{ExecuteAction} dispatches to the workflow appropriate to the selected action's type. Navigation actions invoke a single \textsc{Navigate} call; element actions inside a form section invoke \textsc{SubmitForm}, while other element actions invoke \textsc{ElementAction}, which routes to the appropriate workflow (App.~\ref{app:workflows}); the end-task action invokes \textsc{LLMVerifyEndTask}. If the chosen action raises a runtime error (an invalid URL, a stale selector, an interaction failure on a non-interactable element), the action is removed from $\mathcal{A}_t$ and the LLM is re-prompted; this retry budget resets each timestep and is bounded at three attempts.
 
\paragraph{End-task verification.}
When the LLM selects end-task, \textsc{LLMVerifyEndTask} issues an LLM call conditioned on the task instruction $I$ and the interaction history $h_t$ that judges whether the task has been completed (Prompt~\ref{prompt:VerifyEndTask}). If completion is verified, a separate \textsc{LLMFinalAnswer} call (Prompt~\ref{prompt:FinalAnswer}) produces the answer string and the loop terminates. If not, the end-task action is removed from $\mathcal{A}_t$ for the current timestep only --- it remains available on subsequent timesteps --- and the LLM is re-prompted to choose a different action. We allow at most one verification check per task episode; subsequent end-task actions end the task immediately.
 
\paragraph{Intra-step continuation.}
Some actions (entering input into a field, copying to clipboard) leave the page on the same URL and complete a partial rather than full intent. To avoid the overhead of restarting the observation pipeline for follow-up actions on the same page, after such an action the loop enters a short continuation phase: \textsc{UpdatePageMem} refreshes the page state, a VLM is prompted to describe the visual difference and the description is concatenated with $\hat{o}_t$ to produce the updated observation $\hat{o}'_t$, and the LLM selects another action from a freshly-gathered candidate set. The continuation phase ends when either the page URL changes, a modal dialog appears (handled at the next timestep with focused attention), the agent selects end-task, or a budget of five within-step actions is reached.

\subsection{Action Workflows}
\label{app:workflows}
 
We provide the details of the workflows invoked by \textsc{ExecuteAction} when the agent selects an element action (App.~\ref{app:agent-loop}). Element actions are dispatched in two ways: if the selected element lies inside a form section, control passes to \textsc{SubmitForm}; otherwise it passes to \textsc{ElementAction} (Algorithm~\ref{alg:elementaction}), which routes to the appropriate per-element-type workflow based on tag, role, and DOM attributes. Navigation actions and end-task are handled directly in the agent loop and are not covered here.
 
\paragraph{Element-type dispatch.}
\textsc{ElementAction} is structured as a flat decision tree over element properties: it checks first for behaviors known from exploration (recorded \texttt{dropdown\_elements}), then for input-type-specific handlers (file upload, select/combobox, the various \texttt{input}/\texttt{textarea} subtypes), then for ``probably opens something'' signals (\texttt{aria-haspopup}, or any element not yet explored), and finally falls through to a plain click.
 
\begin{algorithm}[h]
\caption{\textsc{ElementAction}}
\label{alg:elementaction}
\begin{algorithmic}[1]
\Procedure{ElementAction}{Element $e$}
  \If{$e.\text{dropdown\_elements} \neq \emptyset$} \Comment{recorded from exploration}
    \State \Call{DropdownAction}{$e$}
  \ElsIf{$e.\text{input\_type} = \texttt{file}$}
    \State \Call{UploadFile}{$e$}
  \ElsIf{$e.\text{tag} = \texttt{select}\, \lor\, e.\text{role} = \texttt{combobox}$}
    \State \Call{SelectOption}{$e$}
  \ElsIf{$e.\text{tag} \in \{\texttt{input}, \texttt{textarea}\}\, \lor\, e.\text{role} = \texttt{spinbutton}$}
    \If{$e.\text{input\_type} \in \{\texttt{submit}, \texttt{reset}, \texttt{button}\}$}
      \State \Call{ClickElement}{$e$}
    \ElsIf{$e.\text{input\_type} = \texttt{search}$}
      \State \Call{Search}{$e$}
    \ElsIf{$e$ is a radio or checkbox}
      \State \Call{ClickElement}{$e$}
    \Else
      \State \Call{EnterInput}{$e$}
    \EndIf
  \ElsIf{$e.\text{aria-haspopup} \neq \texttt{false}\, \lor\, \lnot\, e.\text{explored}$}
    \State \Call{DropdownAction}{$e$} \Comment{probe for dropdown semantics}
  \ElsIf{$e$ is a copy button}
    \State \Call{CopyToClipboard}{$e$}
  \Else
    \State \Call{ClickElement}{$e$}
  \EndIf
\EndProcedure
\end{algorithmic}
\end{algorithm}
 
\paragraph{Form submission.}
\textsc{SubmitForm} (Algorithm~\ref{alg:submitform}) handles forms in three phases. An LLM call first selects which fields to fill (Prompt~\ref{prompt:SelectFields}), and \textsc{ElementAction} is invoked on each chosen field, dispatching to \textsc{EnterInput}, \textsc{SelectOption}, or \textsc{UploadFile} as appropriate. A validation pass then re-fills any field that is empty-but-required or carries \texttt{aria-invalid="true"} after the initial entry --- the LLM is re-prompted for new values for these fields. Finally, a review loop allows the LLM to inspect the populated form and either edit additional fields, submit, or exit (Prompt~\ref{prompt:SelectFormAction}); exiting leaves the form in its current state and returns control to the main agent loop without submitting. Within the review loop the candidate set is restricted to elements of the form section.
 
\begin{algorithm}[h]
\caption{\textsc{SubmitForm}}
\label{alg:submitform}
\begin{algorithmic}[1]
\Procedure{SubmitForm}{form section $s$}
  \State $F \gets$ \Call{LLMSelectFields}{$s$}
  \For{$f \in F$}
    \State \Call{ElementAction}{$f$} 
    \State \Call{UpdateSection}{$s$}
  \EndFor
  \For{$f \in s.E_s$ where $f$ is empty-and-required \textbf{or} $f.\text{aria-invalid} = \texttt{true}$}
    \State \Call{ElementAction}{$f$}
  \EndFor
  \State \Call{UpdateSection}{$s$}
  \For{$k = 1, \ldots, K_{\max}$} \Comment{review phase, $K_{\max} = 15$}
    \State $a \gets$ \Call{LLMSelectFormAction}{$s$} \Comment{element in $s$, submit, or exit}
    \If{$a$ is exit \textbf{or} $a$ is a submit button}
      \State \textbf{if} $a$ is a submit button \textbf{then} \Call{ClickElement}{$a$}
      \State \textbf{return}
    \EndIf
    \State \Call{ElementAction}{$a$}
    \State \Call{UpdateSection}{$s$}
    \If{current URL has changed} \textbf{return} \EndIf
  \EndFor
\EndProcedure
\end{algorithmic}
\end{algorithm}
 
\paragraph{Dropdown action.}
\textsc{DropdownAction} (Algorithm~\ref{alg:dropdownaction}) clicks the dropdown trigger and consults the section diff returned by \textsc{UpdatePageMem}. Three outcomes are possible. If the URL changed or no new elements were revealed, the click was an ordinary navigation or null action and the workflow returns. If the revealed elements form a coherent form-like cluster (multiple inputs together with a submit-like button), control is routed to \textsc{SubmitForm} on the synthesized form section. Otherwise --- the typical case of a menu, autocomplete list, or option dropdown --- the LLM selects one of the revealed elements (Prompt~\ref{prompt:SelectAction}) and that element is clicked. The form-detection heuristic \textsc{IsForm} returns true when $\Delta^+$ contains at least two input-like elements and at least one element matching submit-button heuristics (a button with type \texttt{submit}, or accessible text matching submit-like keywords).
 
\begin{algorithm}[h]
\caption{\textsc{DropdownAction}}
\label{alg:dropdownaction}
\begin{algorithmic}[1]
\Procedure{DropdownAction}{element $e$}
  \State $u_{\text{pre}} \gets$ current URL
  \State \Call{ClickElement}{$e$}
  \State $\Delta \gets$ \Call{UpdatePageMem}{current page}
  \If{current URL $\neq u_{\text{pre}}$ \textbf{or} $\Delta^+ = \emptyset$} \Return \EndIf
  \If{\Call{IsForm}{$\Delta^+$}}
    \State \Call{SubmitForm}{synthesize form section from $\Delta^+$}
  \Else
    \State $e' \gets$ \Call{LLMSelectAction}{$\Delta^+$}
    \State \Call{ClickElement}{$e'$}
  \EndIf
\EndProcedure
\end{algorithmic}
\end{algorithm}
 
\paragraph{Other workflows.}
\begin{itemize}
    \item \textsc{Search}: invokes \textsc{EnterInput} on the search field, then computes a section diff to detect whether suggestions have appeared. If they have, the LLM is offered the option to select a suggestion (which is then clicked) or to ignore them (Prompt~\ref{prompt:SelectSearchSuggestions}). The workflow concludes by pressing the Enter key to issue the search.
 
    \item \textsc{EnterInput}: prompts the LLM for the value to enter (Prompt~\ref{prompt:EnterInputValue}) and fills it into the field.
 
    \item \textsc{UploadFile}: presents the LLM with the choice of either an existing file in the agent's local filesystem (input files staged for the task --- e.g., the input images supplied with VisualWebArena tasks --- and any text files created earlier in the same task or VLM-captioned images saved during the task) or a \emph{create-new-file} option (Prompt~\ref{prompt:UploadFile}). In the latter case the agent is prompted for a filename and text content (Prompt~\ref{prompt:CreateFile}), the file is written to the local filesystem, and the new file is then uploaded.
 
    \item \textsc{SelectOption}: prompts the LLM to choose one of the available options (Prompt~\ref{prompt:SelectOption}) and sets the field's value to that option via Playwright.
 
    \item \textsc{CopyToClipboard}: reads the copied text from the clipboard and logs it in the action history.
 
    \item \textsc{ClickElement}: issues a Playwright click on the element.
\end{itemize}

\subsubsection{Action Logging and History Format}
\label{app:history-format}

Every successful basic action contributes a string to the agent's interaction history. Failed actions (those that raise the runtime errors handled by the retry mechanism in App.~\ref{app:agent-loop}) are not logged --- only the eventually-successful action appears. Action strings are constructed \emph{after} execution, since several formats reference values that are known only post-execution (e.g., the actual text entered into a field, the option that was selected). Table~\ref{tab:action-strings} lists the format for each basic action.

\begin{table}[h]
\centering
\small
\begin{tabular}{ll}
\toprule
\textbf{Action} & \textbf{Log string format} \\
\midrule
\textsc{ClickElement}      & \texttt{Clicked \{elem\_name\}} \\
\textsc{CopyToClipboard}   & \texttt{Copied text to clipboard: "\{value\}"} \\
\textsc{SelectOption}      & \texttt{Selected value of "\{option\}" for field "\{elem\_name\}"} \\
\textsc{EnterInput}        & \texttt{Entered value of "\{input\_value\}" into \{elem\_name\}} \\
\textsc{UploadFile}        & \texttt{Uploaded file: "\{file\}" into the "\{elem\_name\}" field} \\
\textsc{SwitchTab}         & \texttt{Switched from tab \{start\_i\} (\{start\_url\}) to tab \{new\_i\} (\{new\_url\})} \\
\textsc{SwitchWebsite}     & \texttt{Navigated to the "\{website\_name\}" website (\{website\_url\})} \\
\textsc{GoToPage}          & \texttt{Navigated to page "\{page\_name\}" (\{page\_url\})} \\
\textsc{GoBack}            & \texttt{Returned to the "\{page\_name\}" page} \\
\textsc{EnterURL}          & \texttt{Navigated to URL: "\{url\}"} \\
\bottomrule
\end{tabular}
\caption{Log string format for each basic action.}
\label{tab:action-strings}
\end{table}

Compound actions and intra-step continuation chains produce multiple basic-action strings within a single timestep. These are emitted as a list under an \texttt{Actions:} heading in the history; a timestep that produced exactly one basic action uses the singular \texttt{Action:} heading instead. Each timestep contributes a block of the form below to the history, with the page name and URL drawn from the PageMem, the task summary from \textsc{AnalyzePage}, and the reason produced by the LLM in the same call as the action selection itself.

\begin{verbatim}
* Step 1:
  * Observation: {page_name} ({url})
    - Summary: {task_summary}
  * Reason for Action: {reason}
  * Action: {action_str}
* Step 2:
  * Observation: {page_name} ({url})
    - Summary: {task_summary}
  * Reason for Action: {reason}
  * Actions:
    - {action_str}
    - {action_str}
\end{verbatim}

\section{Additional Experiment Details}
\label{app:experiment-details}

\paragraph{Exploration parameters.}
Offline exploration is bounded by four limits per website: a maximum of 75 clickable elements explored per page, 500 pages per website, and a search depth of 2 (where the homepage is at depth 0, so a depth-2 traversal covers three layers). Each website is also subject to a 12-hour wall-clock timeout, after which exploration terminates and the partial WebsiteMem is used as-is. For Online-Mind2Web, which spans 136 distinct websites, depth is reduced to 1 and the per-website timeout to 1 hour. WorkArena uses the same parameters as WebArena and VisualWebArena.

\paragraph{URL replacement on WebArena and VisualWebArena.}
WebArena and VisualWebArena evaluate against locally-hosted simulated copies of real websites (Reddit, GitLab, OpenStreetMap, etc.), but task instructions refer to these sites by their real names. We observed that LLMs frequently misinterpret this as a directive to navigate to the real site --- e.g., reading the localhost URL, concluding ``I am not on Reddit,'' and attempting to navigate to \url{https://www.reddit.com}, which breaks evaluation. We address this confusion with a bidirectional URL substitution applied at the prompt boundary: simulated-site URLs are rewritten to their real-site counterparts in every string passed to the LLM, and the inverse rewrite is applied to URLs in LLM outputs before they reach the browser. The LLM thus reasons consistently as if it were operating on the real site, while the browser remains pointed at the simulation. Table~\ref{tab:url-replacements} lists the seven substitutions used.

\begin{table}[h]
\centering
\small
\begin{tabular}{ll}
\toprule
Simulated host & Real-site URL \\
\midrule
\texttt{\$SHOPPING}        & \url{https://www.onestopmarket.com} \\
\texttt{\$CLASSIFIEDS}     & \url{https://osclass-classifieds.com} \\
\texttt{\$REDDIT}          & \url{https://www.reddit.com} \\
\texttt{\$WIKIPEDIA}       & \url{https://en.wikipedia.org} \\
\texttt{\$SHOPPING\_ADMIN} & \url{http://magento.site/} \\
\texttt{\$GITLAB}          & \url{https://gitlab.com} \\
\texttt{\$MAP}             & \url{http://openstreetmap.org} \\
\bottomrule
\end{tabular}
\caption{URL substitutions applied at the prompt boundary on WebArena and VisualWebArena. Environment variables hold the localhost URLs of the simulated sites. The substitution is applied bidirectionally: simulated $\to$ real on input to the LLM, real $\to$ simulated on URLs in the LLM's outputs.}
\label{tab:url-replacements}
\end{table}

\paragraph{Multi-website selection (WebArena).}
At the start of each task, the agent is presented with the full list of benchmark websites and prompted to select any sites beyond the starting URL that are relevant (Prompt~\ref{prompt:Multi-site}). The homepages of selected sites are added to the bookmark set $B_\tau$, making them available as one-click navigation actions throughout the task.

\paragraph{Input image grounding (VisualWebArena).}
A subset of VisualWebArena tasks include input images that the agent must reason over alongside the task instruction. At task start, the VLM is prompted with the task instruction, the input image(s), and the current page screenshot, and asked to produce a textual description of the image(s) in relation to the task (Prompt~\ref{prompt:InputImage}). This description is appended to the task instruction for the duration of the task.

\paragraph{Hyperparameters.}
Table~\ref{tab:hyperparameters} consolidates the configuration values used across all components of the system. These hyperparameters were largely chosen heuristically as reasonable defaults and were not extensively swept, as we did not observe strong sensitivity in pilot runs.

\begin{table}[h]
\centering
\small
\begin{tabular}{lll}
\toprule
Component & Parameter & Value \\
\midrule
Exploration & max elements per page & 75 \\
            & max pages per website & 500 \\
            & max depth (homepage at depth 0) & 2 \quad (1 for OM2W) \\
            & max time per website & 12h \quad (1h for OM2W) \\
\midrule
Page division & oversize thresholds $(h, w)$ & $(\!>\!900,\!>\!320)$ or $(\!>\!500,\!>\!800)$ \\
              & list-grouping run length & $\geq 4$ \\
\midrule
Section update & resummarization threshold $|\Delta^+| + |\Delta^-|$ & $\geq 3$ \\
\midrule
Observation & list-item chunk size $c$ & 25 \\
            & minimum image size for VLM description & $50 \times 50$ px \\
\midrule
Agent loop & max steps $T_{\max}$ & 30 \\
           & intra-step continuation budget $J_{\max}$ & 5 \\
           & action-error retries per step & 3 \\
           & end-task verification attempts per task & 1 \\
\midrule
Form submission & review loop bound $K_{\max}$ & 15 \\
\bottomrule
\end{tabular}
\caption{Consolidated hyperparameter values. Per-benchmark exploration overrides are noted parenthetically; all other values are identical across the four benchmarks.}
\label{tab:hyperparameters}
\end{table}

\subsection{Compute Cost Estimates}
\label{app:compute-cost}

Experiments were performed on a desktop machine with Ryzen 5 3600 CPU, NVIDIA RTX 3090 GPU, and 64GB RAM. Inference was run locally using vLLM \citep{kwon2023efficientmemorymanagementlarge}. Total execution time for each benchmark was ${\sim}7$ days for WebArena, ${\sim}8$ days for VisualWebArena, ${\sim}3$ days for Online-Mind2Web, and ${\sim}2$ days for WorkArena. Based on estimated system power draw and regional electricity prices, we estimate that experiments cost roughly \$1.15 in electricity per day, leading to a total estimate of \$23 for the four benchmarks. On average, each task used 270k tokens total across the LLM and VLM, which would translate to roughly \$0.03 per task if using OpenRouter API endpoints at the time of writing. Exploration used approximately 50M total tokens for summarization across all benchmark websites.

\section{Broader Impacts}
\label{app:impacts}
Our work shows that capable web agents can be built on small, locally-runnable open-weight models, which has positive implications for cost, privacy, and research accessibility: automation of tedious web tasks becomes economical at scales where frontier-model APIs would not, sensitive browsing sessions need not leave the user's device, and reproducible agent research becomes more tractable for groups without large compute budgets. However, this also lowers the barrier for misuse such as spam posting, fake account creation, and review manipulation. Agents acting autonomously over long horizons also raise deployment concerns: even the strongest current agents make mistakes, and the compounding effect of errors across multi-step tasks means we recommend human oversight for any consequential domain.

\section{Limitations}
\label{app:limitations}
Our framework relies on hand-designed components that encode structural priors about how web pages are typically organized, such as DOM-based section decomposition, heuristics for identifying clickable elements, deterministic exploration rules, and a fixed set of compound-action workflows. While our implementation is generally robust across a wide range of websites, performance may degrade on sites that diverge significantly from common patterns. Our method also utilizes a larger number of sequential LLM calls, which increases wall-clock time per task and makes the framework expensive to run with frontier models. We further investigate only a minimal instantiation of the memory component; richer mechanisms such as online workflow learning or synthetic-data generation are left to future work. Finally, all of our evaluation is conducted on benign tasks, and the system's robustness to adversarial page content is uncharacterized \citep{tur2025safearenaevaluatingsafetyautonomous, zheng2025webguardbuildinggeneralizableguardrail, xiang2025guardagentsafeguardllmagents, wu2026botsbaitexposingmitigating, ying2026securewebarenaholisticsecurityevaluation, zhang2025browsesafeunderstandingpreventingprompt,  zhang2025attackingvisionlanguagecomputeragents, wu2025dissectingadversarialrobustnessmultimodal, liao2026redteamcuarealisticadversarialtesting, kuntz2025osharmbenchmarkmeasuringsafety, anthropic2025promptinjection}.

 
\section{Prompts}
 
\subsection{Observation Prompts}
 
\begin{promptbox}{SelectSections}
\label{prompt:SelectSections}
You are an intelligent virtual assistant who completes tasks for users on various websites. You will be given information about your task, the previous steps taken, and a list of summaries for the different sections available on the current page. You must now identify all page sections that seem relevant to the task based on their descriptions and select those sections for further analysis.
\\
\\The information will be given as:
\\TASK: task request from the user
\\HISTORY: summary of previous task steps
\\CURRENT PAGE: name and url of current page
\\PAGE SECTIONS: list of sections available on the current page
\\
\\Think step-by-step and identify all page sections that might contain relevant details or potentially aid in progressing the task. Then, select the relevant page sections by providing a list of integers corresponding to the indices of your selected sections.
\\
\\Give your full thought process behind your reasoning as:
\\**THOUGHTS** your reasoning steps
\\
\\Give your answer for the relevant sections as a comma-separated sequence of integers corresponding to your choices:
\\**RELEVANT SECTIONS**: indices of selected page sections (e.g., "2, 5, 3")
\\
\\Additional Guidelines:
\\- Select all sections that seem like they could potentially be useful at any point in the task.
\\- If you think a section might be relevant but are not sure, select it to see its full details.
\end{promptbox}
 
\begin{promptbox}{ExtractDetails}
\label{prompt:ExtractDetails}
You are an intelligent virtual assistant who completes tasks for users on various websites. To complete your task, you must first analyze each section of the page one-by-one and identify any relevant details. You will be given information about your overall task and the next section of the page to analyze. You must then summarize any information or elements in the page section that might potentially be relevant for the task.
\\
\\The following information will be provided:
\\TASK: your task instructions
\\HISTORY: summary of previous task steps
\\CURRENT PAGE: title, url, and summary of current page
\\SECTION: page section to analyze
\\CONTENT: section content
\\
\\First, think step-by-step about the current subsection of the page you are analyzing and consider whether more information needs to be gathered. Then, provide a summary of the information and save the relevant details in this format: `- **\{descriptive\_label\}**: "\{value\}"`
\\
\\Give your full thought process followed by your answer in this format:
\\**THOUGHT**: your reasoning
\\**SUMMARY**: summary of why the information may or may not be relevant to the task
\\**RELEVANT DETAILS**: noteworthy information from this section
\\
\\Additional guidelines:
\\- Do not include instructions for the next steps to take just yet, only highlight information or potentially useful actions from the current page.
\\- Never make assertive statements or assumptions about the given page section as it only provides a partial view of the website. Always mention any potential uncertainties and/or alternative possibilities about the information given.
\\- Make sure to mention any page details that could help us find the correct path to take even if the information is not directly relevant itself.
\\- Your response should end after you provide the relevant details.
\end{promptbox}
 
\begin{promptbox}{SummarizePage}
\label{prompt:SummarizePage}
You are a virtual AI assistant who performs tasks on behalf of users through a web browser. You will be provided with your assigned task instructions, the previous interaction history, and the current web page observation. You must now write a concise summary of the current page that highlights the relevant information for your task.
\\
\\The following information will be provided:
\\TASK: your task instructions
\\HISTORY: previous task steps
\\CURRENT PAGE: title and url of current page
\\PAGE OBSERVATION: overview of page
\\
\\First, think step-by-step about the progress you have made previously in order to understand the context of the current situation. Then, identify the information on the page that is the most relevant to your task and provide a concise report (max one paragraph in length) that summarizes the current observation.
\\
\\Provide your full thought process followed by your summary of the page observation in this format:
\\**THOUGHT**: your reasoning steps
\\**OBSERVATION SUMMARY**: summary of relevant details
\\
\\Additional Guidelines:
\\- You should not decide what action to take yet, only highlight relevant information.
\\- Do not make assumptions about the functionality of the elements on the page. You may describe what you think will be a likely result of clicking an action, but you should mention any potential uncertainties and not make strong statements.
\\- Make sure to include in your summary any details from the current page that would be worth remembering in the future.
\\- If the page is not relevant to your task then simply provide a brief description of the page and why it is not relevant.
\\- Your answer should end after you provide the observation summary.
\end{promptbox}
 
\subsubsection{List item selection prompts.}
\begin{promptbox}{SelectItems}
\label{prompt:SelectItems}
You are an intelligent virtual assistant who completes tasks for users on various websites. You will be requested by the user to complete a task that involves finding one or more items in a list of content (e.g. a list of links, products, articles, comments, etc.) on the website. The list of items on the current page will be shown and you will have the option to select any items that are relevant to the task based on their positional indices.
\\
\\The following information will be provided:
\\TASK: the user's task instructions
\\HISTORY: summary of previous task steps
\\PAGE: title and url of the current page
\\LIST INFO: summary describing the content list
\\LIST ITEMS: the list items on the current page of results
\\
\\Think step-by-step about the given information and determine if there are any relevant list items on the current page of results. Then, select all items that are relevant to the task in any way by providing a comma-separated list of integers corresponding to the positional indices of those items (e.g. "2, 9, 13"). If there are no exact matches in the list, then still select the item that is the closest match. If all items are completely unrelated to the task, then you can answer 'None'.
\\
\\Provide the full thought process behind your answer followed by your final answer for the item selection in this format:
\\
\\**THOUGHTS**: your reasoning steps
\\
\\**SELECT ITEMS**: list items to select (list of integers or 'None')
\\
\\Additional Guidelines:
\\- If some of the list items seem like potential matches but there is not enough information to confirm, select them in order to view more details.
\\- Select all items that are partially related to the task requirements even if they are not exact matches.
\\- Always include all items that are mentioned in the task instructions.
\\- End your response after selecting the list items.
\end{promptbox}
 
\begin{promptbox}{CheckDone}
\label{prompt:CheckDone}
You are an intelligent virtual assistant who completes tasks for users on various websites. Your current objective requires you to find one or more items in a list of content (e.g. a list of pages, products, articles, comments, etc.). In the previous steps you have iterated over the items in order and recorded any matching items found. You will be given the history of the items you have checked so far and you must determine if the objective is complete or if you still need to find more items in the list.
\\
\\You will be provided with the following information:
\\PAGE: title, url, and description of the current page
\\OBJECTIVE: the task you must complete
\\ITEMS CHECKED: list of items that have been checked so far, and whether they match
\\
\\Think step-by-step about the requirements of the objective and the details about the items that have been checked so far. Then, determine if the task is complete or if you still need to look for more items in the rest of the list.
\\
\\Give your full thought process and your answer in this format:
\\**THOUGHTS**: steps behind your reasoning process
\\**COMPLETE**: whether the objective has been completed (Yes/No)
\\
\\Reminder:
\\- Pay attention to the sort order of the list. It may allow you to determine if the remaining list items are worth checking (e.g. if you need to find the cheapest item and the list is sorted by price, then the remaining items will be more expensive).
\\- If it is still possible that there might be more matching items in the list, then you should keep searching to be sure.
\end{promptbox}

\subsection{Action Prompts}
 
\subsubsection{Agent Loop Prompts}
\begin{promptbox}{SelectBookmarks}
\label{prompt:SelectBookmarks}
You are a virtual AI assistant who performs tasks on behalf of users through a web browser. You will be given a user task request and a list of pages on the current website that you can visit in order to carry out the task. You must now identify any pages that are likely to be relevant for completing the assigned task.
\\
\\The list of available pages will be given in this format:
\\ALL PAGES:
\\	1) \{page\_1\}
\\	2) \{page\_2\}
\\	...
\\	n) \{page\_n\}
\\
\\You should carefully think step-by-step to determine whether any of the pages would likely be useful for completing the task. Include all pages that have the potential to be helpful for completing the task in any way. Provide your reasoning as:
\\**THOUGHTS**: your reasoning steps
\\
\\Finally, select the relevant pages by providing a comma-separated sequence of integers corresponding to the indices of your choices:
\\**RELEVANT PAGES**: indices of selected pages(s) (e.g. "1, 7")
\\
\\Note:
\\- Select any pages that seem like they might be partially relevant or help find the target page quicker.
\\- If there are no pages that seem relevant for the task then provide 'None' as your answer.
\end{promptbox}
 
\begin{promptbox}{Element candidates}
\label{prompt:ElementCandidates}
You are a virtual AI assistant who performs tasks on behalf of users through a web browser. You will be provided with a list of clickable elements available on the current page and you must identify all candidate elements that are potentially useful for progressing the task.
\\
\\Your assigned task and the previous interaction history will be provided as:
\\TASK: task instructions
\\HISTORY: summary of previous task steps
\\CURRENT PAGE: title, url, and summary of current page observation
\\
\\The elements present on the page will be listed in this format:
\\PAGE ELEMENTS:
\\	1) \{element\_1\}
\\	2) \{element\_2\}
\\	...
\\
\\You should carefully think step-by-step to determine which elements might help progress the task. Provide your full reasoning process as:
\\**THOUGHTS**: your reasoning steps
\\
\\Finally, select one or more elements by providing the indices of your choices separated by commas:
\\**SELECT ELEMENTS**: indices of selected element(s) (e.g. "1, 7")
\\
\\Note:
\\- For search/filtering interfaces, make sure to clear all pre-existing filters before applying a new one.
\\- If there are multiple elements that seem relevant and you are not sure which one is best, then include all relevant elements in your answer.
\\- If none of the listed elements are relevant then answer 'None'.
\end{promptbox}
 
\begin{promptbox}{Navigation candidates}
\label{prompt:NavigationCandidates}
You are a virtual AI assistant who performs tasks on behalf of users through a web browser. You will be provided with your assigned task instructions, the previous interaction history, current page observation, and a list of browser navigation actions to choose from. You must now determine if any of the available navigation actions would be useful for progressing the task.
\\
\\The following information will be provided:
\\TASK: your task instructions
\\HISTORY: summary of previous task steps
\\CURRENT PAGE: title, url, and summary of current page
\\BROWSER ACTIONS: list of available navigation options and their details
\\
\\Think step-by-step about the navigation options and whether they would be helpful or harmful for progressing the task. Then, select one or more candidates for the next action to take. If none of the navigation options are useful for the task then select 'None'.
\\
\\Give your full thought process behind your reasoning as:
\\**THOUGHTS** your reasoning steps
\\
\\Select one or more of the navigation options by providing a comma-separated sequence of integers corresponding to the indices of your choices:
\\**ANSWER**: indices of selected option(s) (e.g. "1, 7")
\end{promptbox}
 
\begin{promptbox}{SelectAction}
\label{prompt:SelectAction}
You are an autonomous AI assistant who performs tasks on behalf of users through a web browser. You will be provided with your assigned task instructions, the previous interaction history, current page observation, and a list of valid actions to choose from. You must now determine the correct next action to take in order to progress with the task, or end the task if it has been completed.
\\
\\The following information will be provided:
\\TASK: your task instructions
\\HISTORY: summary of previous task steps
\\CURRENT PAGE: title, url, and summary of current page observation
\\ACTIONS: list of available action options
\\
\\You should first carefully think step-by-step about the potential benefits and risks of each option in order to determine the logical next action for making progress on your task. Provide your full thinking process as:
\\**THOUGHTS**: your reasoning steps
\\
\\Once you have decided on the next action you wish to take, provide the reason behind your decision followed by the index of your selected action in the following format:
\\**REASON**: reason for decision
\\**SELECT ACTION**: index of selected action (e.g. "1")
\\
\\Note:
\\- The reason in your final answer should be a one sentence summary of the thinking process behind your decision.
\\- You must select exactly ONE action out of the options provided.
\\- Do not mark the task as complete until you have fully completed all steps specified by the instructions (e.g. if the task is to buy a product, you need to complete the full checkout and payment process).
\\- Only provide the index of the selected action with no additional text afterwards.
\end{promptbox}
 
\begin{promptbox}{VerifyEndTask}
\label{prompt:VerifyEndTask}
You are an intelligent virtual assistant who completes tasks for users on various websites. Based on your given task instructions, your previous action history and the current page observation, you must determine whether you have fully completed the task.
\\
\\The following information will be provided:
\\TASK: task instructions from the user
\\HISTORY: summary of previous task steps
\\CURRENT PAGE: name, url and summary of the current browser page
\\ACTIONS: available actions (not executed yet)
\\
\\Think step-by-step about user's task instructions and identify which parts of the task you have accomplished, then determine whether the task is fully complete or additional actions need to be taken.
\\
\\**Guidelines for evaluation**:
\\1. You are only required to complete the steps explicitly defined by the instructions.
\\2. As long as there are any next steps that need to be taken, the task completion status should be false.
\\3. Tasks that involve buying an item should only be considered complete after adding the item to the cart and completing the full checkout process.
\\4. If the user asks you to show them an item (e.g. "show me the most expensive ...", "show me the most recent ...", "find me a ..." etc.), then the task is not complete until the full details of the item are displayed by clicking its link.
\\5. If the task instructions asks you to open a page (e.g. "open my latest ..."), then the task is not fully complete until you have clicked the link for the page.
\\
\\Provide your thought process followed by your answer in this format:
\\**THOUGHTS**: your reasoning process
\\**TASK COMPLETE**: task completion status (True/False)
\end{promptbox}
 
\begin{promptbox}{FinalAnswer}
\label{prompt:FinalAnswer}
You are an intelligent virtual assistant who performs tasks for users on various websites. Previously, you have completed the user's task on the website and now you must send a final message to the user.
\\
\\You will be provided with the following information:
\\USER: the task request from the user
\\HISTORY: summary of previous task steps
\\CURRENT PAGE: title, url, and summary of current page
\\
\\Think step-by-step the user's task instructions and the information you have found on the webstite. Then, provide a message to the user that summarizes the steps you have taken to complete their task followed by the final answer to the user's query.
\\
\\Outline your full thought process as:
\\**THOUGHTS**: your reasoning
\\
\\Provide your message to the user followed by your answer value in this format:
\\**MESSAGE**: completion message
\\**ANSWER**: final answer value
\\
\\Important guidelines for the final answer:
\\- If you were unable to successfully complete the task or no answer is necessary, provide "N/A" as the answer.
\\- When asked to return a count, return the count as a number with units instead of "N/A" if it's 0.
\\- If the user asks you to check if something is true or not, answer "Yes" or "No".
\end{promptbox}
 
\subsubsection{Action Workflow Prompts}
\begin{promptbox}{SelectFields}
\label{prompt:SelectFields}
You are an intelligent virtual assistant who completes tasks for users on various websites. You will be requested by the user to complete a task that involves filling in a web form. The list of available form fields will be provided and you will be able to select which ones to edit based on the task.
\\
\\The task instructions from the user and your current subgoal will be provided as:
\\TASK: the task assigned by the user
\\HISTORY: the previous interaction history
\\CURRENT PAGE: name, url and summary the of current page
\\
\\The form will be provided as a numbered list of available input fields:
\\FORM:
\\1) \{field\}
\\2) \{field\}
\\...
\\n) \{field\}
\\
\\Think carefully about the task and identify the input fields that you need to select, then provide your reasoning as:
\\**THOUGHT**: {reasoning}
\\
\\Finally, provide the list the of all form fields that should be updated for the task by providing a comma-separated sequence of integers corresponding to the indices of your choices as:
\\**EDIT FIELDS**: {list of indices} (e.g. "1, 7")
\end{promptbox}
 
\begin{promptbox}{SelectFormAction}
\label{prompt:SelectFormAction}
You are an intelligent virtual assistant who completes online form filling tasks on various websites. In the previous steps, you filled out the available fields of the form and you will now choose the next action to take in the form.
\\
\\The task information will be given in the following format:
\\TASK: overall task to complete
\\PROGRESS: the current task progress
\\FORM MENU: numbered list of actions available in the form section
\\
\\Think step-by-step about the given information and identitify the appropriate next action to take among those listed in the form menu. You will then be able to select one of the actions by providing its index.
\\
\\Provide your full thought process followed by your action choice as:
\\**THOUGHTS**: your reasoning steps
\\**CHOICE**: index of selected action (e.g. "**CHOICE**: 14")
\\
\\Additional Guidelines:
\\- Don't repeat the action performed in the previous step.
\\- End your response after providing your action choice.
\end{promptbox}
 
\begin{promptbox}{Enter Input Value}
\label{prompt:EnterInputValue}
You are an intelligent virtual assistant who completes tasks for users on various websites. You will be provided information about your assigned task, the web page, and the current state of the task progress. You have just selected an input field on the page and you can now provide a value to enter into the input field.
\\
\\The task information will be provided in this format:
\\
\\TASK: <user's task instructions>
\\
\\PAGE: <summary of page>
\\
\\PROGRESS: <current task progress>
\\
\\INPUT FIELD: <selected input field>
\\
\\Think step-by-step about the actions needed to complete the task and then answer with the value to enter into the selected input field. If the task specifies an exact input value to enter then your answer should match. Otherwise, come up with a reasonable input value to enter given the task context.
\\
\\Provide your full thought process followed by your final answer for the input value as:
\\
\\**THOUGHTS**: <reasoning>
\\
\\**INPUT VALUE**: <value>
\\
\\Additional Guidelines:
\\- If the task does not directly specify an input value to use then think carefully about what would be the best value to enter in the field.
\\- Only provide an input value for the current selected input field with no additional text afterwards.
\end{promptbox}
 
\begin{promptbox}{Select Search Suggestions}
\label{prompt:SelectSearchSuggestions}
You are an intelligent virtual assistant who completes tasks on various websites. You will be provided information about the current page and your assigned task. A search bar displaying a list of search suggestions is currently selected and you will now be have the option to click one of them.
\\
\\The information about your task, the current browser page and the selected search bar will be provided as:
\\TASK: <your assigned task>
\\PAGE: <summary of page>
\\SEARCH FIELD: <details about the search bar>
\\SUGGESTIONS: <list of search suggestions>
\\
\\Think step-by-step about whether any of the search suggestions are relevant to the task. If you want to click one of the suggestions then provide the number and value of the option. If there are no relevant search suggestions then you can choose the 'None' option. If the correct input value is already entered into the field and the same value also appears in the search suggestions, you should still select the same value again.
\\
\\Provide your full thought process followed by your final answer in this format:
\\**THOUGHTS**: <reasoning process>
\\**SELECT**: <number and value of option> (e.g. "1) Meta")
\end{promptbox}
 
\begin{promptbox}{Select Option}
\label{prompt:SelectOption}
You are an intelligent virtual assistant who completes tasks on various websites. You will be provided information about the current page and your assigned task. You have selected an input field on the page with several options available and you must select the appropriate numbered option from the list by providing its index.
\\
\\The information about the page, your task, and the options for the input field will be provided in this format:
\\
\\TASK: task instructions
\\
\\PROGRESS: current task progress
\\
\\PAGE: summary of page
\\
\\INPUT FIELD: selected input element
\\
\\INPUT OPTIONS:
\\	1) \{option\_1\}
\\	2) \{option\_2\}
\\	...
\\	n) \{option\_n\}
\\
\\Think step-by-step about the actions needed to complete the task and then answer with the value to enter into the selected input field. If the task specifies an exact input option to select then your answer should match. Otherwise, choose the option that most closely matches with the task instructions given the task context.
\\
\\Provide your full thought process followed by your final answer as:
\\
\\**THOUGHTS**: your reasoning
\\
\\**OPTION**: index of selected option (e.g. "1")
\\
\\Additional Guidelines:
\\- Only select ONE of the input options.
\\- Don't provide any additional text after the selected option.
\end{promptbox}
 
\begin{promptbox}{Upload File}
\label{prompt:UploadFile}
You are an intelligent virtual assistant who completes tasks for users on various websites. Your current task involves uploading a file and you have already navigated to the part of the website where the file upload should take place. The list of files available in your filesystem will be given and you must choose now choose the appropriate file to upload.
\\
\\The following information will be provided:
\\TASK: your task instructions
\\HISTORY: summary of the previous actions you have taken
\\CURRENT PAGE: title, url, and summary of current page
\\INPUT FIELD: the field that you are uploading the file to
\\FILES: list of files to choose from
\\
\\Based on this information, you should carefully analyze the context of your current task and determine which file from your filesystem should be uploaded. If the task requires creating a new file to upload, then you can select the 'Create new file' option to write and upload a file.
\\
\\Think step-by-step and provide your full thought process as:
\\**THOUGHTS**: your reasoning steps
\\
\\Then, choose one of the options by providing the index of your choice as:
\\**ANSWER**: chosen index (e.g. "1")
\end{promptbox}
 
\begin{promptbox}{Create File}
\label{prompt:CreateFile}
You are an intelligent virtual assistant who completes tasks for users on various websites. Your current task involves creating a new file and uploading it to the current website. Based on the given task context, you must create the file that will be uploaded by providing the name and contents of the file.
\\
\\The following information will be provided:
\\TASK: your task instructions
\\HISTORY: summary of the previous actions you have taken
\\CURRENT PAGE: title, url, and summary of current page
\\FIELD: the field where the file will be uploaded
\\
\\Reason carefully about your task and plan what you will name the file and what the file contents should be. Once you are ready, you can provide the file name followed by the full file content and this file will be uploaded to the website.
\\
\\Provide your full thinking process as:
\\**THOUGHTS**: your reasoning steps
\\
\\Provide the name of the file followed the file content in this format:
\\
\\**FILE**: file name (e.g. urls.txt)
\\
\\**CONTENT**:
\\```
\\\# full file content here ...
\\```
\end{promptbox}

\begin{promptbox}{Enter URL}
\label{prompt:EnterURL}
You are an autonomous virtual assistant who completes tasks for users on various websites by controlling a web browser. You will be provided information about your assigned task, the previous actions you have taken, and the current page information. The browser address bar has been selected and you must now provide a URL to enter and navigate to.
\\
\\You will be given the following information:
\\TASK: your task instructions
\\HISTORY: summary of previous task steps
\\CURRENT PAGE: title, url, and summary of current page observation
\\
\\Think carefully about the given task information and determine the correct URL for the page that you wish to navigate to on the website.
\\
\\Important guidelines:
\\1. The URL must be for the same website as the current page.
\\2. Pay close attention to the previous URLs you have visited in order to help understand the URL structure used by the website.
\\
\\Provide your full reasoning process followed by your answer for the URL in this format:
\\**THOUGHTS**: your reasoning steps
\\**ANSWER**: URL
\end{promptbox}

\subsection{Other Prompts}
 
\subsubsection{WebArena}
\begin{promptbox}{Multi-site}
\label{prompt:Multi-site}
You are an autonomous AI assistant who performs tasks on behalf of users through a web browser. You will be given a task from the user and start on a website where the task needs to be completed on. First you must determine if the task only involves the current website, or if part of the task needs to be completed on another website.
\\
\\You will be given the following information:
\\USER: task instructions from the user
\\CURRENT PAGE: description of current browser page
\\WEBSITES: list of available websites
\\
\\Carefully read the user's instructions and reason about whether the task involves only the current website, or if a task requirements involves another website. In most cases the task will only focus on the current site but occasionally it will require information or functionality from one of the other websites. Your final answer should include one or more websites that are necessary for the task and must always include the current website. 
\\
\\Provide your reasoning process followed by your final answer as:
\\**THOUGHTS**: your reasoning steps
\\**ANSWER**: website(s) needed for the task (e.g "OneStopMarket", "Reddit + GitLab")
\end{promptbox}
 
\subsubsection{VisualWebArena}
\begin{promptbox}{Input Image}
\label{prompt:InputImage}
Describe what the following user is asking for based on their request message that references both the first image and the web page screenshot.
\\
\\**User request message**: "\{task\_instruction\}"
\end{promptbox}

\subsubsection{VLM Prompts}
\begin{promptbox}{VLM Summarize Section}
\label{prompt:VLMSummarizeSection}
Describe this section of the \{website\_name\} website in one sentence.
\end{promptbox}

\begin{promptbox}{VLM Summarize Page}
\label{prompt:VLMSummarizePage}
Describe this page of the \{website\_name\} website in one sentence.
\end{promptbox}

\begin{promptbox}{VLM Screen Diff}
\label{prompt:VLMScreenDiff}
These two screenshots show the \{website\_name\} website immediately before and after I \{action\_str\}. What was the result of the action?
\end{promptbox}

\begin{promptbox}{Describe Image - Short}
\label{prompt:DescImageShort}
Describe the image with a short sentence.
\end{promptbox}

\begin{promptbox}{Describe Image - Long}
\label{prompt:DescImageLong}
Fully extract all information contained in this image in an organized format.
\\
\\Make sure to include:
\\- All textual and/or numerical information.
\\- Visual features such as objects, colors, and settings.
\\- Names of any recognizable people, celebrities, characters, media, or locations.
\end{promptbox}


\newpage

\end{document}